\documentclass[11pt,a4paper]{article}
\usepackage{times,latexsym}
\usepackage{url}
\usepackage[T1]{fontenc}
\usepackage{amsmath}
\usepackage{amsfonts}
\usepackage{booktabs}
\usepackage[table]{xcolor}
\usepackage{subcaption}
\usepackage{multirow}

\usepackage[acceptedWithA]{tacl2021v1}
\usepackage{graphicx}
\usepackage{xspace,mfirstuc,tabulary}
\usepackage{subcaption}

\newif\iftaclinstructions
\taclinstructionsfalse 
\iftaclinstructions
\renewcommand{\confidential}{}
\renewcommand{\anonsubtext}{(No author info supplied here, for consistency with
TACL-submission anonymization requirements)}
\newcommand{\instr}
\fi

\iftaclpubformat 

\else

\fi

\title{One Model to Translate Them All? A Journey to Mount Doom for Multilingual Model Merging} 

\author{
  Baban Gain\thanks{\hspace{0.5em}Corresponding authors.} \quad
  Trilok Nath Singh \quad
  Asif Ekbal\footnotemark[1]
  \\
  Indian Institute of Technology Patna, India \\
  \texttt{\{gainbaban,tnsiitb,asif.ekbal\}@gmail.com}
}

\date{}

\begin{document}
\maketitle
\begin{abstract}
Weight-space model merging combines independently fine-tuned checkpoints without access to the original training data. While merging has shown promise in multitask settings, its behavior in multilingual generative systems remains underexplored. We systematically study weight-space merging for multilingual machine translation by fully fine-tuning language models on large-scale bilingual corpora and evaluating representative merging strategies across shared-source, shared-target, and bidirectional consolidation settings. Our experiments reveal a strong directional asymmetry. Merging is comparatively more effective when models share a target language, improving multilingual coverage over the base model, but it still fails to preserve the peak performance of language-specific checkpoints. In contrast, when target languages differ, performance degrades sharply, especially in shared-source and bidirectional settings. To explain this behavior, we analyze internal representations and find that fine-tuning does not create disjoint language-specific sub-networks. Instead, independently fine-tuned models activate largely overlapping neurons while reshaping upper-layer target-generation representations into incompatible geometries. These findings suggest that multilingual merging failures arise from target-side geometric misalignment within shared computational units, challenging the assumptions underlying standard weight-space merging for multilingual translation. We make the code publicly available at \url{https://github.com/babangain/mt-model-merging}.

\end{abstract}

\section{Introduction}

Large language models (LLMs) have shown remarkable progress across a range of tasks, such as machine translation \cite{gain2026bridginglinguisticdividesurvey}, summarization \cite{ZHANG2026131928}, reasoning \cite{bandyopadhyay2025thinkingmachinessurveyllm}, code generation \cite{10.1145/3747588}, etc. While many modern systems are designed to support multiple languages, training a single multilingual model that performs well across diverse language pairs remains computationally expensive and data-intensive \cite{aharoni-etal-2019-massively,he2024scalinglawsmultilinguallanguage,seto-etal-2025-assessing}. Further, when multilingual corpora are highly imbalanced, naive mixing or data-proportional sampling can lead to poor performance on low-resource languages, while oversampling the low-resource languages can degrade high-resource performance \cite{arivazhagan2019massivelymultilingualneuralmachine,chang-etal-2024-multilinguality}. A common alternative is to fine-tune smaller monolingual models independently \cite{10.1007/978-3-031-78172-81_7,raihan-zampieri-2025-tigerllm}, which is more feasible for low-resource or domain-specific settings but results in many specialized models that are costly to host, maintain, and deploy at scale \cite{10.1145/3787849}.

Consolidating independently fine-tuned models is difficult because training data is often unavailable due to privacy or licensing constraints, making joint retraining unrealistic. Model merging addresses this by combining weights directly without data, and has been known to be effective across multiple tasks. However, its behavior in multilingual settings remains underexplored for fully fine-tuned bilingual generative MT systems.

Recently, some works have combined merging with continued pre-training to inject low-resource or code-mixed capabilities into a base model before downstream fine-tuning \cite{tao-etal-2024-unlocking,kodali2025adaptingmultilingualmodelscodemixed}, where the goal is improved adaptation or classification performance rather than consolidation of fully specialized generative systems. Domain-focused studies further consider merging general and domain-specific models to enhance terminology retention across languages \cite{rousset2025merginglanguagedomainspecific}, targeting vocabulary acquisition instead of end-to-end generative alignment. In contrast, our setup merges independently fine-tuned, full-parameter bilingual machine translation models trained on million-scale corpora.


In this paper, we systematically study model merging in multilingual settings using machine translation as a controlled testbed. Machine Translation naturally involves distinct source and target languages, allowing us to probe disparities between understanding and generation in LLMs. We fine-tune models on Indic–English pairs (Hindi, Bengali, Tamil, Telugu $\leftrightarrow$ English), which share a pivot language while differing in typological properties. We evaluate merging across three configurations: shared source language, shared target language, and merging unidirectional models to form bidirectional systems, enabling analysis of how language configuration affects weight-space fusion.

Our results show that multilingual merging behaves differently from standard multitask merging and introduces unique challenges.  
Our contributions can be summarized as follows:
\begin{itemize}
    \item To the best of our knowledge, this is one of the first systematic studies of weight-space merging for multilingual MT using full-parameter fine-tuning on million-scale bilingual corpora, with checkpoints released for benchmarking.
    \item Demonstration of strong directional asymmetry and larger degradation than multitask merging, especially when target languages differ or when forming bidirectional systems.
    \item Span based activation analysis shows that fine-tuning redistributes language specialization to upper layers and creates geometric misalignment that harms merging. Restoring target-language-specific upper layers improves translation performance, providing preliminary support for routing-based architectures.
\end{itemize}

\section{Related Work}

\subsection{Model Merging}

Early work on model soups \citep{wortsman2022model} showed that simple weight averaging across fine-tuned checkpoints can improve robustness without additional training. Task Arithmetic \citep{ilharco2022editing} formalized this idea through \textit{task vectors}, which are defined as the parameter difference between a fine-tuned model and its pretrained initialization. By linearly combining such vectors and adding them to the base model, multiple task capabilities can be composed post hoc.
The naive addition of task vectors, however, introduces parameter interference when tasks induce conflicting updates. TIES \citep{10.5555/3666122.3666432} addresses this by pruning low-magnitude updates and resolving sign conflicts before merging, thereby reducing destructive interference. DARE \citep{3692070.3694452} instead randomly drops a large fraction of parameters in the delta vectors and rescales the remainder to preserve the expected magnitude, encouraging sparsity prior to fusion. Fisher merging \citep{matena2022merging} weights parameters according to estimated sensitivity and performs a Fisher-weighted average to emphasize directions that are important for each task.
More recently, geometric approaches have been proposed. Subspace Boosting \citep{skorobogat2025subspaceboostedmodelmerging} and TSV-Merging \citep{gargiulo2025tasksingularvectorsreducing} decompose task vectors via singular value decomposition and restrict merging to dominant subspaces. By suppressing low-variance components, these methods aim to preserve coherent task-relevant structure while limiting interference.

Despite this growing body of work, most evaluations have focused on vision backbones or English-centric multitask benchmarks \citep{huang2024emrmerging,qi2024moreefficientmodelmerging}. Recent systematic analyses indicate that techniques effective in vision do not reliably transfer to large language models \citep{hitit2025systematicstudymodelmerging}. In particular, merging behavior under multilingual specialization remains underexplored. When independently fine-tuned models specialize in different languages, representational shifts may be deeper and structurally asymmetric. Our work investigates this setting directly.

\subsection{Neurons in LLMs}

Understanding multilingual specialization requires examining neuron-level behavior. \citet{geva-etal-2021-transformer} characterized feed-forward layers in Transformers as key--value memories, where individual neurons store contextual associations retrieved during generation. ROME \citep{meng2022locating} later provided causal evidence for localized factual knowledge by identifying and editing small subsets of neurons responsible for such knowledge in MLP layers.
Subsequent studies have analyzed activation patterns and selectivity. \citet{voita-etal-2024-neurons} observed that many neurons remain largely inactive across inputs, while others exhibit highly selective responses to specific tokens or short n-grams. In multilingual models, \citet{tang-etal-2024-language} introduced language activation metrics showing that small neuron subsets disproportionately contribute to language-specific processing. \citet{mondal-etal-2025-language} further explored whether manipulating such language-specific neurons through test-time activation replacement or LoRA updates restricted to selected units could improve cross-lingual transfer, though gains were inconsistent.

While prior work emphasizes localization, sparsity, and interventions within a single model, our study examines how neuron-level specialization evolves under independent fine-tuning and how it interacts with weight-space merging.

\section{Methodology}






\begin{table*}[!ht]
\centering
\rowcolors{2}{gray!15}{white}
\resizebox{\textwidth}{!}{%
\begin{tabular}{@{}l*{10}{c}@{}}
\toprule
& \multicolumn{5}{c}{\textbf{BLEU}} & \multicolumn{5}{c}{\textbf{CHRF}} \\
\textbf{Model} & \textbf{Hindi} & \textbf{Bengali} & \textbf{Tamil} & \textbf{Telugu} & \textbf{Average} & \textbf{Hindi} & \textbf{Bengali} & \textbf{Tamil} & \textbf{Telugu} & \textbf{Average} \\
\midrule
Base & 18.75 & 14.97 & 3.96 & 5.69 & 10.84 & 43.29 & 39.08 & 21.26 & 24.76 & 32.10 \\
Multilingual f/t: Indic$\rightarrow$En & 28.67 & 20.73 & 12.37 & 9.64 & 17.85 & 52.96 & 45.87 & 36.57 & 30.96 & 41.59 \\ 
Multilingual f/t: Indic$\leftrightarrow$En & 24.63 & 16.64 & 7.94 & 5.67 & 13.72 & 51.92 & 44.51 & 31.92 & 27.57 & 38.98   \\ \midrule

Finetuned - Hindi & \textbf{37.77} & 17.93 & 0.86 & 2.39 & 14.74 & \textbf{62.21} & 45.43 & 15.36 & 23.01 & 36.50 \\
Finetuned - Bengali & 25.13 & \textbf{32.99} & 1.23 & 3.87 & 15.80 & 50.95 & \textbf{59.42} & 16.26 & 25.30 & 37.98 \\
Finetuned - Tamil & 18.22 & 12.17 & \textbf{27.97} & 2.95 & 15.33 & 42.92 & 36.94 & \textbf{54.72} & 22.15 & 39.18 \\
Finetuned - Telugu & 19.00 & 11.19 & 0.67 & \textbf{31.77} & 15.66 & 43.98 & 35.80 & 12.90 & \textbf{57.61} & 37.57 \\
\midrule
\multicolumn{11}{c}{\textbf{\textit{Merged Models}}} \\
\midrule
Task Arithmetic & 30.97 & 26.51 & 10.71 & 17.20 & 21.35 & 56.29 & 52.80 & 34.06 & 42.66 & 46.45 \\
TIES & 32.44 & 26.91 & 10.65 & 18.34 & 22.08 & 58.23 & 53.67 & 35.21 & 44.69 & 47.95 \\
DARE & 31.22 & 26.71 & 11.02 & 18.87 & 21.96 & 56.93 & 53.05 & 34.49 & 43.81 & 47.07 \\
SCE-Merging & 33.47 & 27.02 & 9.98 & 18.05 & 22.13 & 59.42 & 54.37 & 35.23 & 45.11 & 48.53 \\
\bottomrule
\end{tabular}%
}
\caption{BLEU and CHRF scores for different models across languages in the Indic$\rightarrow$English direction.}
\label{tab:results-indic-en}
\end{table*}

\begin{table*}[!ht]
\centering
\rowcolors{2}{gray!15}{white}
\resizebox{\textwidth}{!}{%
\begin{tabular}{@{}l*{10}{c}@{}}
\toprule
 & \multicolumn{5}{c}{\textbf{BLEU}} & \multicolumn{5}{c}{\textbf{CHRF}} \\
\textbf{Model} & \textbf{Hindi} & \textbf{Bengali} & \textbf{Tamil} & \textbf{Telugu} & \textbf{Average} & \textbf{Hindi} & \textbf{Bengali} & \textbf{Tamil} & \textbf{Telugu} & \textbf{Average} \\
\midrule
Base & 5.93 & 1.82 & 0.92 & 0.60 & 2.32 & 27.97 & 23.54 & 24.12 & 15.78 & 22.85 \\
Multilingual f/t: En$\rightarrow$Indic & 23.59 & 10.75 & 5.05 & 5.25 & 11.16 & 49.59 & 43.71 & 35.28 & 28.75 & 39.33 \\
Multilingual f/t: En$\leftrightarrow$Indic & 20.80 & 9.19 & 4.33 & 4.37 & 9.67 &46.71 & 41.36 & 34.05 & 26.66 & 37.19  \\

\midrule
Finetuned - Hindi & \textbf{29.23} & 0.32 & 0.41 & 0.40 & 7.59 & \textbf{54.47} & 0.57 & 0.47 & 0.56 & 14.02 \\
Finetuned - Bengali & 0.06 & \textbf{15.45} & 0.12 & 0.18 & 3.95 & 0.23 & \textbf{50.13} & 0.15 & 0.20 & 12.68 \\
Finetuned - Tamil & 0.35 & 0.39 & \textbf{12.61} & 0.96 & 3.58 & 0.64 & 0.43 & \textbf{51.75} & 0.76 & 13.39 \\
Finetuned - Telugu & 0.45 & 0.44 & 0.79 & \textbf{15.59} & 4.32 & 0.71 & 0.44 & 0.64 & \textbf{50.17} & 12.99 \\
\midrule
\multicolumn{11}{c}{\textbf{\textit{Merged Models}}} \\
\midrule
Task Arithmetic & 6.42 & 2.15 & 0.45 & 0.54 & 2.39 & 28.38 & 23.89 & 17.20 & 12.46 & 20.48 \\
TIES & 0.73 & 0.09 & 0.10 & 0.09 & 0.25 & 12.56 & 3.75 & 3.10 & 0.70 & 5.03 \\
DARE & 1.38 & 0.35 & 0.22 & 0.15 & 0.52 & 16.55 & 11.18 & 7.85 & 3.83 & 9.85 \\
SCE-Merging & 9.88 & 3.34 & 0.74 & 0.50 & 3.61
& 34.08 & 28.46 & 19.23 & 13.19 & 23.74 \\\bottomrule
\end{tabular}%
}
\caption{BLEU and CHRF scores for different models across languages in the English$\rightarrow$Indic direction.}
\label{tab:results-en-indic}
\end{table*}
\subsection{Large-Scale MT Fine-tuning}
We fine-tune the base \textbf{Qwen-2.5-3B-Instruct} model independently on eight bilingual translation tasks covering four language pairs from the \textbf{Samanantar} corpus~\cite{ramesh-etal-2022-samanantar}: English--Hindi, English--Bengali, English--Tamil, and English--Telugu. Fine-tuning is performed using the \textbf{LLaMAFactory} framework~\cite{zheng-etal-2024-llamafactory} with full-parameter updates that allow the entire model to specialize for each language pair.

The selected language pairs are intentionally diverse. Hindi and Bengali belong to the Indo-Aryan language family, whereas Tamil and Telugu are Dravidian languages, enabling cross-family analysis. The training sets are substantial: 10.1M sentence pairs for English--Hindi, 8.6M for English--Bengali, 5.26M for English--Tamil, and 4.95M for English--Telugu, for each translation direction. Given the large-scale supervision and full-parameter updating, the resulting models constitute strong task-specialized systems. We therefore treat these independently fine-tuned checkpoints as practical upper bounds for their respective language pairs. They serve as competitive references when evaluating merging-based multilingual consolidation and are subsequently used as inputs to our model merging experiments.

\subsection{Baselines}
\label{sec:baselines}

We compare against several representative weight-space merging methods, along with the pretrained backbone and multilingual fine-tuning as references.

\textbf{Pretrained Model.} This is the original instruction-tuned backbone prior to task-specific fine-tuning. It serves as a lower bound and quantifies the gains obtained through specialization and merging.

\textbf{Multilingual Joint Fine-tuning.}
We train multilingual reference models by pooling data across related directions: (a) English$\rightarrow$four Indic languages, (b) four Indic languages$\rightarrow$English, and (c) all eight directions jointly. This follows the broader joint multilingual MT paradigm, where a single model is optimized over multiple translation directions, while our setup remains English-pivot rather than fully many-to-many \cite{JMLR:v22:20-1307}. These models provide context for the behavior of joint multilingual supervision when all training data and budget are available. However, they are not directly comparable to weight-space merging: model merging assumes a post-training setting in which independently fine-tuned checkpoints already exist, and the practitioner may not have the data access, compute budget, or incentive to train a new fully multilingual model from scratch.

\textbf{Task Arithmetic} \citep{ilharco2022editing}. Fine-tuning is modeled as a task vector $\Delta_t = \theta_t - \theta_0$, where $\theta_0$ is the pretrained backbone. Multiple tasks are merged via linear combination, $\theta_{\text{merged}} = \theta_0 + \sum_i \alpha_i \Delta_i$. Although simple, direct addition may introduce parameter interference when updates conflict.

\textbf{TIES} \citep{10.5555/3666122.3666432}. TIES reduces destructive interference by pruning low-magnitude updates, resolving sign conflicts for each parameter, and merging the filtered deltas through normalized summation.

\textbf{DARE} \citep{3692070.3694452}. DARE sparsifies task vectors by randomly dropping a proportion $p$ of parameters and rescaling the remainder by $\frac{1}{1-p}$. The sparsified deltas are then merged using standard weight-space techniques.

\textbf{SCE-Merging} \citep{wan-etal-2025-fusechat}. SCE selects significant weight changes relative to a pivot model, assigns layer-wise merging weights based on update strength, removes conflicting directions, and integrates the resulting deltas into the pivot.

\section{Experimental Findings}

We evaluate weight-space merging under three controlled multilingual configurations: (i) shared target language (Many$\rightarrow$One), (ii) shared source language (One$\rightarrow$Many), and (iii) bidirectional construction through merging directionally opposite models. 

\subsection{Common Target Language: Many$\rightarrow$One}

We merge independently fine-tuned Indic$\rightarrow$English models that share a fixed target language (English) but differ in their source languages (Hindi, Bengali, Tamil, Telugu), isolating multilingual source aggregation within a constant generation space.

As shown in Table~\ref{tab:results-indic-en}, individual fine-tuned models perform strongly on their supervised pairs but generalize poorly to unseen sources. Merging reduces this imbalance: CHRF scores become consistently distributed across languages, with merged models reaching 46--48 on average, compared to 32.10 for the base model. Merged checkpoints also exceed both the average cross-lingual performance of any single fine-tuned model and the performance of the multilingual models. However, peak task performance is not preserved. The gap from the fine-tuned upper bounds is larger than typically reported in multitask merging \cite{he2025mergebenchbenchmarkmergingdomainspecialized}. Thus, while Many$\rightarrow$One merging improves multilingual coverage and stability, it remains suboptimal for retaining maximum task accuracy.
\begin{table*}[h!]
\centering
\rowcolors{2}{gray!15}{white}
\resizebox{\textwidth}{!}{%
\begin{tabular}{lcccccccccc}
\toprule
 & \multicolumn{5}{c}{\textbf{En$\rightarrow$XX}} & \multicolumn{5}{c}{\textbf{XX$\rightarrow$En}} \\
\textbf{Model} & \textbf{Hindi} & \textbf{Bengali} & \textbf{Tamil} & \textbf{Telugu} & \textbf{Average} & \textbf{Hindi} & \textbf{Bengali} & \textbf{Tamil} & \textbf{Telugu} & \textbf{Average} \\
\midrule
Base & 5.93 & 1.82 & 0.92 & 0.60 & 2.32 & 18.75 & 14.97 & 3.96 & 5.69 & 10.84 \\
Multilingual f/t: En$\leftrightarrow$Indic & 20.80 & 9.19 & 4.33 & 4.37 & 9.67 & 24.63 & 16.64 & 7.94 & 5.67 & 13.72 \\
Fine-tuned & 29.23 & 15.45 & 12.61 & 15.59 & - & 37.77 & 32.99 & 27.97 & 31.77 & - \\

\midrule
\multicolumn{11}{c}{\textbf{\textit{Merged Models}}} \\
\midrule
Task Arithmetic & 6.23 & 2.16 & 1.05 & 1.29 & 2.68 & 29.49 & 24.33 & 16.35 & 22.82 & 23.25 \\
TIES & 0.58 & 0.44 & 1.04 & 1.30 & 0.84 & 29.27 & 21.24 & 20.05 & 23.07 & 23.41 \\
DARE & 0.59 & 0.47 & 1.00 & 1.26 & 0.83 & 30.09 & 24.92 & 19.39 & 22.71 & 24.28 \\
SCE-Merging & 10.20 & 6.56 & 0.99 & 1.25 & 4.75 & 28.03 & 22.12 & 18.10 & 21.27 & 22.38 \\

\bottomrule
\end{tabular}%
}
\caption{BLEU scores for bidirectional models obtained by merging English$\rightarrow$XX and XX$\rightarrow$English models.}
\label{tab:bidirectional-bleu}
\end{table*}

\subsection{Common Source Language: One$\rightarrow$Many}

We next consider the complementary configuration where models share a common source language (English) but differ in their target languages (Table~\ref{tab:results-en-indic}). 
Both BLEU and CHRF show that target-language generation expertise is not effectively aggregated. Average performance, in many cases, falls below the pretrained baseline, and degradation relative to fine-tuned upper bounds is substantially more severe than in the Many$\rightarrow$One case. This level of retention is far below the 70--90\% commonly observed in multitask merging \cite{he2025mergebenchbenchmarkmergingdomainspecialized,3692070.3694452}, indicating that combining distinct target-side generation spaces is considerably more destructive than aggregating diverse source encoders under a shared target. Multilingual models perform much better than merged models; however, their performance remains significantly lower than that of individually fine-tuned models. This is consistent with observations from \citet{sachan-neubig-2018-parameter}, where multilingual training can degrade accuracy when target languages belong to distant language families.

\subsection{Bidirectional Construction via Opposite Directions}

We finally evaluate whether merging can construct a bidirectional model by combining English$\rightarrow\ell$ and $\ell\rightarrow$English checkpoints for each $\ell \in \{\mathrm{hi}, \mathrm{bn}, \mathrm{ta}, \mathrm{te}\}$, as shown in Table~\ref{tab:bidirectional-bleu}.

Although the merged scores may appear decent at first glance, it is important to note that each result corresponds to merging only two checkpoints: one English$\rightarrow\ell$ model and one $\ell\rightarrow$English model. Thus, the setting is considerably simpler than the multilingual merging experiments, yet the merged models still remain far from the corresponding direction-specific upper bounds.

For English$\rightarrow$Indic, BLEU retention is around 35\% for Hindi, 42\% for Bengali, and below 10\% for Tamil and Telugu. In contrast, Indic$\rightarrow$English is partially preserved, although BLEU still drops relative to the upper bound. We report CHRF scores in \autoref{appendix:chrf-birdirectional}, which show severe degradation. Here, the multilingual fine-tuned models (row 2 in \autoref{tab:bidirectional-bleu}) are much weaker because they have been fine-tuned on eight language pairs, compared to the single-direction fine-tuned models in the third row.
Overall, merging opposite translation directions causes substantially greater degradation than unidirectional merging, even when only two models are merged.

\section{Analysis}
\subsection{Language-Specific Neuron Behavior Before and After Fine-tuning}
To quantify language specialization at the neuron level, we analyze span-conditioned MLP gate activations measured during forward pass. Our objective is to isolate source-side and target-side behavior within a single autoregressive sequence and to identify neurons that are both selective and strongly activated for particular languages. Our methodology is inspired by the LAPE framework \cite{tang-etal-2024-language}, which motivates representation-level analysis for understanding language-specific specialization within multilingual models.

\paragraph{Sequence construction.}
For each language $\ell \in \mathcal{L}$, let
$\mathcal{D}_\ell = \{(u_i^{(\ell)}, t_i^{(\ell)})\}_{i=1}^{N_\ell}$
denote translation examples, where $u_i^{(\ell)}$ contains a source-language sentence embedded in an instruction prompt and $t_i^{(\ell)}$ is the corresponding target-language sentence. Each pair is rendered into a single token sequence
$\mathbf{z}_i^{(\ell)} = (z_{i1}, \dots, z_{in})$.
Conceptually, the sequence decomposes as
\[
\mathbf{z}_i^{(\ell)} =
(\underbrace{p_{i1},\dots,p_{ir_i}}_{\text{instruction}},
\underbrace{x_{i1},\dots,x_{iS_i}}_{\text{source}},
\underbrace{y_{i1},\dots,y_{iT_i}}_{\text{target}}).
\]
Instruction tokens are excluded from all subsequent measurements.

\paragraph{Span masks.}
We define two disjoint masks over token positions:
$m^{\mathrm{src}}_{ij} = \mathbf{1}[z_{ij}\in\{x_{i1},\dots,x_{iS_i}\}]$
and
$m^{\mathrm{tgt}}_{ij} = \mathbf{1}[z_{ij}\in\{y_{i1},\dots,y_{iT_i}\}]$.

Because the model is decoder-only with causal attention, the source-span activations are independent of target tokens, whereas target-span activations are conditioned on the entire source segment.

\paragraph{Span-conditioned activation rates.}
Let $L$ denote the number of layers and $I$ the intermediate MLP width. Let $G^{(l)}_{ijk}$ denote the post-nonlinearity gate activation of neuron $k$ at position $j$ in layer $l$. For span $s \in \{\mathrm{src}, \mathrm{tgt}\}$, define the positive activation count
\[
C_{l,k}^{(\ell,s)} =
\sum_{i=1}^{N_\ell} \sum_{j=1}^{n}
\mathbf{1}[G^{(l)}_{ijk} > 0] \, m_{ij}^{(s)},
\]
and the total number of masked tokens
$N^{(\ell,s)} = \sum_{i=1}^{N_\ell} \sum_{j=1}^{n} m_{ij}^{(s)}$.
The empirical activation probability is then
\(
p_{l,k}^{(\ell,s)} = \frac{C_{l,k}^{(\ell,s)}}{N^{(\ell,s)}}.
\)
The quantity $p_{l,k}^{(\ell,s)}$ represents the probability that neuron $(l,k)$ produces a positive gate activation when processing span $s$ under language $\ell$.

\paragraph{Cross-language selectivity.}
For each neuron $(l,k)$, we normalize activation rates across languages as
$q_{l,k}^{(\ell,s)} =
p_{l,k}^{(\ell,s)} / \sum_{\ell' \in \mathcal{L}} p_{l,k}^{(\ell',s)}$,
with $\sum_{\ell} q_{l,k}^{(\ell,s)} = 1$.
We quantify language selectivity via entropy
\(
H_{l,k}^{(s)} =
- \sum_{\ell \in \mathcal{L}}
q_{l,k}^{(\ell,s)} \log q_{l,k}^{(\ell,s)}.
\)
Low entropy indicates that activation is concentrated in a small subset of languages, whereas high entropy indicates shared multilingual behavior. We select the fraction $\rho$ of neurons with the lowest entropy, i.e.,
$\mathcal{S}^{(s)} = \operatorname*{arg\,min}_{(l,k)}^{\lfloor \rho L I \rfloor} H_{l,k}^{(s)}$.

\paragraph{High-activation criterion.}
Relative selectivity does not guarantee that a neuron is strongly engaged. To ensure that selected neurons exhibit substantial activation, we impose a global activation threshold.
Let
$\mathcal{P} = \{ p_{l,k}^{(\ell,s)} : \forall l,k,\ell \}$
denote the collection of all span-conditioned activation probabilities. We define a threshold $\tau$ as a high percentile of $\mathcal{P}$. 
A selected neuron $(l,k) \in \mathcal{S}^{(s)}$ is assigned to language $\ell$ only if
$p_{l,k}^{(\ell,s)} > \tau$.
This generates per-layer neuron index sets
\[
\mathcal{I}_l^{(\ell,s)} =
\{\, k : (l,k)\in\mathcal{S}^{(s)} 
\text{ and } p_{l,k}^{(\ell,s)} > \tau \,\}.
\]

\paragraph{Interpretation.}

\begin{table}[t]
\centering
\small
\resizebox{\columnwidth}{!}{
\begin{tabular}{lcccc}
\toprule
& \multicolumn{2}{c}{\textbf{Indic $\rightarrow$ En}} 
& \multicolumn{2}{c}{\textbf{En $\rightarrow$ Indic}} \\
\cmidrule(lr){2-3} \cmidrule(lr){4-5}
\textbf{Language} 
& \textbf{Source (src)} & \textbf{Target (tgt)}
& \textbf{Source (src)} & \textbf{Target (tgt)} \\
\midrule
\textbf{Hindi}   
& 542 $\rightarrow$ 1004 
& 1680 $\rightarrow$ 3472
& 706 $\rightarrow$ 958
& 697 $\rightarrow$ 743 \\

\textbf{Bengali} 
& 701 $\rightarrow$ 827 
& 1693 $\rightarrow$ 3460
& 686 $\rightarrow$ 328
& 665 $\rightarrow$ 780 \\

\textbf{Tamil}   
& 950 $\rightarrow$ 2562 
& 2286 $\rightarrow$ 3444
& 662 $\rightarrow$ 516
& 557 $\rightarrow$ 1370 \\

\textbf{Telugu}  
& 546 $\rightarrow$ 1094 
& 2238 $\rightarrow$ 3553
& 534 $\rightarrow$ 1498
& 574 $\rightarrow$ 706 \\
\bottomrule
\end{tabular}
}
\caption{Language-specific total selected neuron counts, summed over all layers. Each cell shows Instruct $\rightarrow$ Fine-tuned counts for the corresponding language and span.}
\label{tab:lang-neuron-counts}
\end{table}

For $s=\mathrm{src}$, the selected neuron sets characterize source-language 
processing, whereas for $s=\mathrm{tgt}$ they characterize generation behavior 
conditioned on the source segment. 
Table~\ref{tab:lang-neuron-counts} reports total selected neuron counts 
(summed over all layers) for both spans and both translation directions.

On the source side, a clear directional asymmetry emerges. 
In Indic$\rightarrow$En, source-span totals increase consistently across 
languages (e.g., Hindi: $542 \rightarrow 1004$, Tamil: $950 \rightarrow 2562$). 
However, this increase is concentrated primarily near the embedding layer 
(Layer~0), while intermediate transformer layers remain largely shared. 
This indicates that fine-tuning mainly strengthens lexical encoding for 
Indic input tokens rather than globally expanding specialization across depth.

The complementary En$\rightarrow$Indic results display a different pattern. 
When English is the source language, source-span totals change modestly and 
may even decrease (e.g., Bengali: $686 \rightarrow 328$). We observe that 
embedding amplification is limited and that mid-layer overlap is largely 
preserved. A layer-wise representation of this behavior is provided in 
Section~\ref{sec:layerwise-heatmap}. Since English representations are already 
well supported in the pretrained model, fine-tuning does not require substantial 
restructuring of source-side encoding.

In contrast, generation-side specialization is substantially stronger. 
Target-span totals increase sharply in Indic$\rightarrow$En 
(e.g., Hindi: $1680 \rightarrow 3472$, Telugu: $2238 \rightarrow 3553$). 
We find that these increases are concentrated in upper transformer layers 
(approximately layers 28--35), which directly govern autoregressive token 
prediction. This indicates deeper reorganization of the generation subspace. 
Importantly, these structural patterns are not restricted to the supervised 
language. Even when a fine-tuned model is probed on a language for which it was 
not explicitly trained, the activation profile often exhibits a qualitatively 
similar depth-dependent structure, particularly in upper layers. This suggests 
that fine-tuning redistributes representational geometry in a way that affects 
multiple languages, not solely the supervised pair. We analyze this 
cross-language generalization behavior more systematically in the next 
subsection. In short, because weight-space merging assumes a degree of geometric 
compatibility across checkpoints, heterogeneous restructuring in these late 
decoding layers leads to misaligned generation subspaces. This asymmetry 
provides a plausible structural explanation for the fragility of weight-space 
merging in multilingual translation.

\subsection{Neuron Usage Alignment}
To examine whether multilingual fine-tuning induces structural separation or shared specialization, we introduce \emph{Neuron Usage Alignment} (NUA). The goal is to determine whether independently fine-tuned bilingual models rely on distinct neuron subsets or modify largely overlapping computational units.

For a model $M$, let $C^{(M,\ell,s)}_{l,k}$ denote the masked positive activation count of neuron $(l,k)$ for language $\ell$ and span $s \in \{\mathrm{src},\mathrm{tgt}\}$, and let $p^{(M,\ell,s)}_{l,k}$ be the corresponding activation rate. We define the layer-wise neuron usage vector
$u^{(M,\ell,s)}_l \in \mathbb{R}^I$
whose entries are these activation rates across the $I$ intermediate neurons. NUA between two models $M_a$ and $M_b$ at layer $l$ is then computed as
\[
\mathrm{NUA}^{(\ell,s)}_l(M_a, M_b)
=
\frac{
u^{(M_a,\ell,s)}_l \cdot u^{(M_b,\ell,s)}_l
}{
\|u^{(M_a,\ell,s)}_l\|_2
\,
\|u^{(M_b,\ell,s)}_l\|_2
}.
\]

High NUA indicates that two models activate largely the same neurons with similar frequencies for the masked span, while lower values indicate divergence in neuron utilization. 
\begin{figure}
\centering
\includegraphics[width=\linewidth]{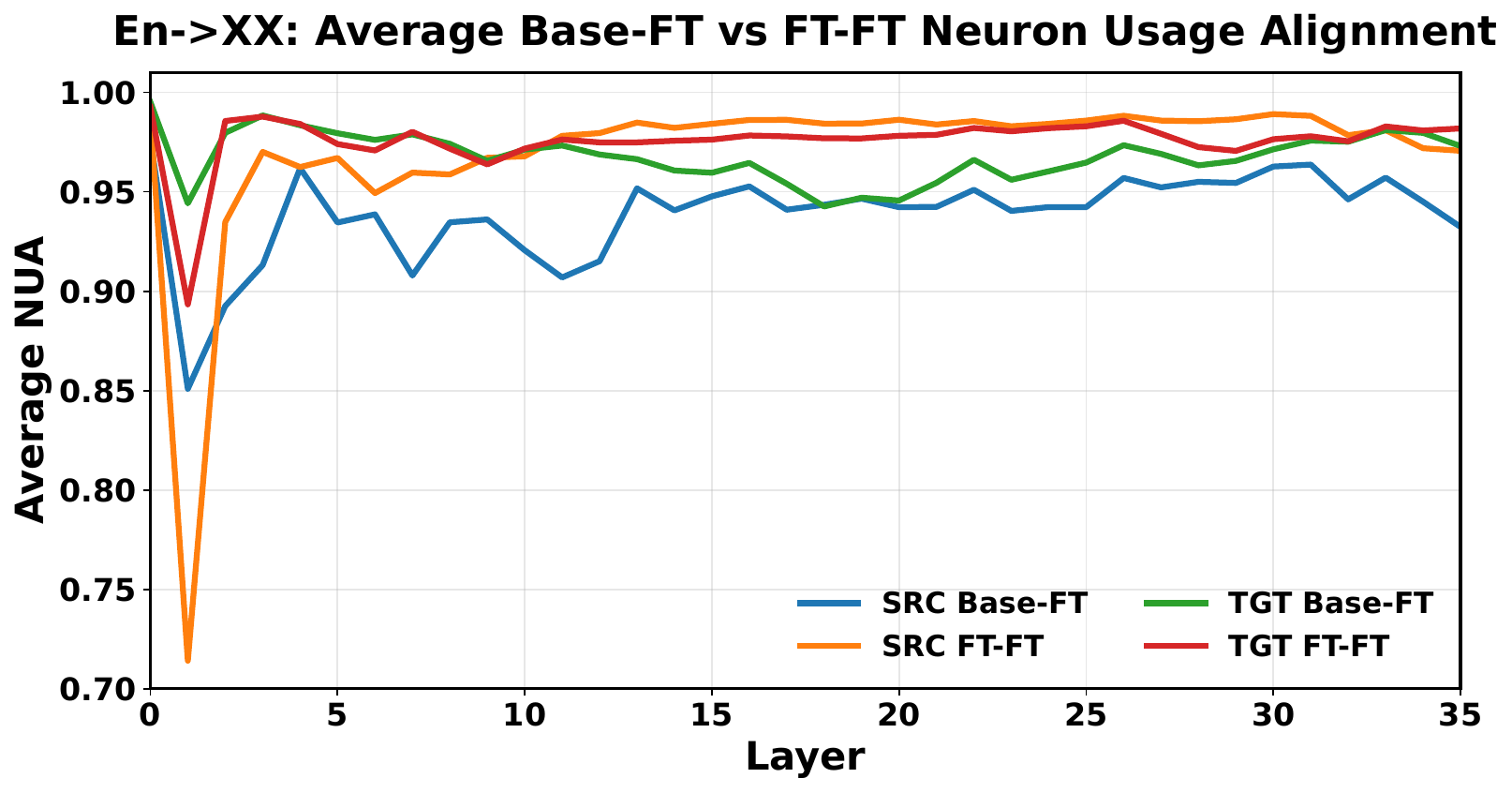}

\includegraphics[width=\linewidth]{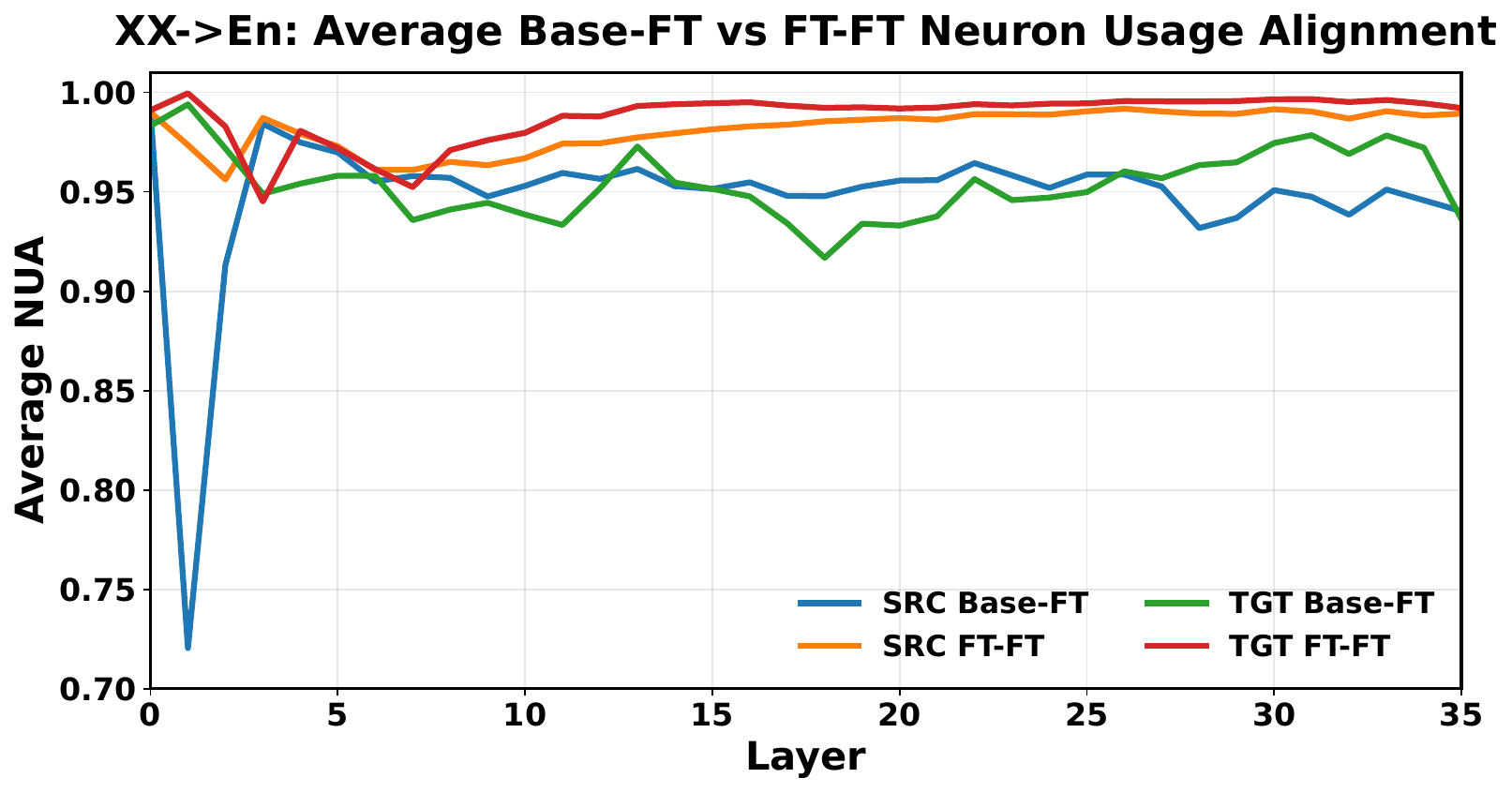}

\caption{Neuron Usage Alignment averages for both translation directions. Top: English$\rightarrow$XX. Bottom: XX$\rightarrow$English. Base-FT denotes average alignment between the base model and fine-tuned models. FT-FT denotes average alignment between each fine-tuned model and the other fine-tuned models.}
\label{fig:nua-avg-both-directions}
\end{figure}

\paragraph{Observations.}
Across both source and target spans, NUA remains consistently high across independently fine-tuned bilingual models. As shown in \autoref{fig:nua-avg-both-directions}, mid and upper layers exhibit near-perfect cosine similarity (typically $>0.98$), indicating that the same intermediate MLP neurons are used with comparable relative frequencies across language pairs. This pattern holds especially strongly in the target span, where autoregressive generation relies on highly overlapping sets of neurons across fine-tuned checkpoints.

Comparisons between the base instruction-tuned model and fine-tuned models show a moderate but systematic reduction in NUA, suggesting that fine-tuning reshapes neuron usage relative to the pretrained baseline. However, the alignment among fine-tuned models themselves remains substantially higher than the alignment between fine-tuned models and the base model. This indicates that bilingual fine-tuning does not allocate disjoint subnetworks for different languages but instead modifies a largely shared set of computational units.

Taken together, these findings suggest that multilingual specialization does not appear to emerge primarily through neuron partitioning. Rather, fine-tuned models engage similar neurons during both encoding and generation, implying that merging degradation is unlikely to arise from structural separation of neuron subsets. Instead, incompatibility likely originates from finer-grained differences within shared units, such as divergent weight directions or representational geometry.

\begin{table*}
\centering
\resizebox{\linewidth}{!}{
\begin{tabular}{l ccc ccc ccc ccc}
\toprule
\multirow{2}{*}{\textbf{Direction}} &
\multicolumn{3}{c}{\textbf{Base vs ft: (src span)}} &
\multicolumn{3}{c}{\textbf{Base vs ft: (tgt span)}} &
\multicolumn{3}{c}{\textbf{Avg other lang: (src span)}} &
\multicolumn{3}{c}{\textbf{Avg other lang: (tgt span)}} \\
\cmidrule(lr){2-4}\cmidrule(lr){5-7}\cmidrule(lr){8-10}\cmidrule(lr){11-13}
&
\textbf{L34} & \textbf{L35} & \textbf{L36} &
\textbf{L34} & \textbf{L35} & \textbf{L36} &
\textbf{L34} & \textbf{L35} & \textbf{L36} &
\textbf{L34} & \textbf{L35} & \textbf{L36} \\
\midrule
En$\rightarrow$Hindi   & 0.7158 & 0.7886 & 0.5972 & 0.9395 & 0.8844 & 0.2721 & 0.9778 & 0.9800 & 0.9524 & 0.9270 & 0.8870 & 0.2486 \\
En$\rightarrow$Bengali & 0.6990 & 0.7796 & 0.5855 & 0.9285 & 0.8773 & 0.2857 & 0.9748 & 0.9808 & 0.9621 & 0.9100 & 0.8719 & 0.2512 \\
En$\rightarrow$Tamil   & 0.7148 & 0.7924 & 0.6051 & 0.8998 & 0.8475 & 0.1482 & 0.9713 & 0.9795 & 0.9658 & 0.7338 & 0.6947 & 0.1187 \\
En$\rightarrow$Telugu  & 0.7216 & 0.7993 & 0.5954 & 0.8911 & 0.8461 & 0.1461 & 0.9739 & 0.9800 & 0.9677 & 0.8073 & 0.7624 & 0.0989 \\
\midrule
Hindi$\rightarrow$En   & 0.7845 & 0.7474 & 0.5493 & 0.9322 & 0.9159 & 0.7897 & 0.8816 & 0.8741 & 0.6203 & 0.9701 & 0.9640 & 0.8863 \\
Bengali$\rightarrow$En & 0.6041 & 0.6355 & 0.4299 & 0.9423 & 0.9332 & 0.8229 & 0.8655 & 0.8662 & 0.6027 & 0.9694 & 0.9651 & 0.8818 \\
Tamil$\rightarrow$En   & 0.6272 & 0.6148 & 0.4704 & 0.9018 & 0.8879 & 0.7061 & 0.6135 & 0.6357 & 0.4310 & 0.9229 & 0.9195 & 0.7782 \\
Telugu$\rightarrow$En  & 0.6361 & 0.6071 & 0.4836 & 0.9190 & 0.9078 & 0.6876 & 0.7459 & 0.7409 & 0.4827 & 0.9552 & 0.9489 & 0.8292 \\
\bottomrule
\end{tabular}
}
\caption{Late-layer linear CKA for layers 34--36. We report (i) same-language checkpoint versus the base model, and (ii) average CKA between the language-specific checkpoint and the other three fine-tuned checkpoints in the same direction.}
\label{tab:cka_late_layers_merged_l34_36}
\end{table*}

\begin{table}
\centering
\small
\resizebox{\columnwidth}{!}{
\begin{tabular}{llcccccc}
\toprule
& & \multicolumn{2}{c}{\textbf{Early (0–11)}} 
  & \multicolumn{2}{c}{\textbf{Mid (12–27)}} 
  & \multicolumn{2}{c}{\textbf{Late (28–36)}} \\
\cmidrule(lr){3-4} \cmidrule(lr){5-6} \cmidrule(lr){7-8}
\textbf{Direction} & \textbf{Span} 
& \textbf{I–FT} & \textbf{FT–FT} 
& \textbf{I–FT} & \textbf{FT–FT} 
& \textbf{I–FT} & \textbf{FT–FT} \\
\midrule
\textbf{Indic$\rightarrow$En} & src 
& 0.992 & 0.995 
& 0.957 & 0.982 
& 0.74 & 0.88 \\

\textbf{Indic$\rightarrow$En} & tgt 
& 0.991 & 0.994 
& 0.952 & 0.978 
& 0.71 & 0.86 \\

\textbf{En$\rightarrow$Indic} & src 
& 0.988 & 0.993 
& 0.938 & 0.972 
& 0.69 & 0.83 \\

\textbf{En$\rightarrow$Indic} & tgt 
& 0.985 & 0.990 
& 0.921 & 0.965 
& 0.64 & 0.80 \\
\bottomrule
\end{tabular}
}
\caption{Layer-banded masked CKA averages. I-FT denotes alignment between the pretrained Instruct model and each fine-tuned checkpoint. FT-FT denotes pairwise alignment among fine-tuned checkpoints. Early, mid, and late correspond to layers 0--11, 12--27, and 28--36, respectively.}
\label{tab:cka_summary}
\end{table}

\subsection{Centered Kernel Alignment}
To quantify how fine-tuning alters internal representations, we employ 
centered kernel alignment (CKA) as a layer-wise measure of 
representational similarity~\cite{10.1007/11564089_7,pmlr-v97-kornblith19a}. 
CKA evaluates whether two models induce similar pairwise similarity structures 
over the same set of inputs.

Let $\{x_i^{(\ell)}\}_{i=1}^{N_\ell}$ denote evaluation inputs for the language 
pair or direction indexed by $\ell$. For transformer layer $l$, let 
$H_l^{(m,\ell)} \in \mathbb{R}^{N_\ell \times d}$ denote the mean-pooled 
hidden representations produced by model $m$ on those inputs. 
For a representation matrix $H$, define the Gram matrix $K = H H^\top$. 
After centering with 
$C = I_{N_\ell} - \frac{1}{N_\ell}\mathbf{1}\mathbf{1}^\top$, 
where $\mathbf{1}$ is the all-ones vector in $\mathbb{R}^{N_\ell}$, 
we compute $\tilde{K} = C K C$. The linear CKA similarity between two 
representation matrices $H_a$ and $H_b$ is
\[
\mathrm{CKA}(H_a, H_b)
=
\frac{\| H_a^\top H_b \|_F^2}
{\| H_a^\top H_a \|_F \, \| H_b^\top H_b \|_F}.
\]

We compute CKA at the same layer index $l$ under two comparisons.

\textbf{Base vs fine-tuned.} 
\(
\mathrm{CKA}\!\left(
H_l^{(\text{base},\ell)},
H_l^{(\text{ft},\ell)}
\right),
\)
which measures how much the geometry at depth $l$ shifts after fine-tuning.

\textbf{Fine-tuned vs fine-tuned.}
\[
\mathrm{CKA}\!\left(
H_l^{(\text{ft},\ell)},
H_l^{(\text{ft},\ell')}
\right),
\qquad \ell \neq \ell',
\]
which measures cross-language geometric alignment across independently 
fine-tuned checkpoints.

Here, $l$ indexes depth and $\ell$ indexes the language pair or direction. 
High CKA indicates preserved geometry in base-to-fine-tuned comparisons or 
shared geometry in cross-checkpoint comparisons, whereas low values reflect 
language-specific reorganization that may undermine weight-space compatibility.

\paragraph{Observations.}

Linear CKA exhibits a clear depth-dependent pattern, but the fine-grained late-layer results in Table~\ref{tab:cka_late_layers_merged_l34_36} reveal that upper-layer divergence is far more structured than layer-band averages alone suggest. In both translation directions, early layers remain almost perfectly aligned with the pretrained model, confirming that lexical encoding and shallow compositional structure are largely preserved after fine-tuning. Middle layers show moderate drift yet retain strong cross-model similarity. The decisive shift occurs in the final decoding block. In the English$\rightarrow$Indic direction, source-span alignment with the base model at layers 34 and 35 remains moderate, while target-span alignment collapses sharply at layer 36, reaching values as low as 0.15--0.27 for several languages. The same pattern appears when comparing the fine-tuned checkpoint for one language with the fine-tuned checkpoints for other languages: source-side cross-language alignment remains extremely high even at layer 36, often above 0.95, but target-side cross-language alignment drops dramatically, in some cases below 0.20. This indicates that English encoding geometry remains mutually compatible across checkpoints, whereas the target-language generation subspace becomes highly language-specific and geometrically misaligned at the final layer.

In contrast, the Indic$\rightarrow$English direction shows substantially greater stability in the late decoding layers for the target span. Even at layer 36, alignment between fine-tuned checkpoints and the base model remains relatively high for English generation, and cross-language alignment among fine-tuned checkpoints frequently exceeds 0.85. Although some source-side divergence appears, the shared English generation space remains structurally coherent across independently fine-tuned models. The late-layer shifts are non-monotonic; for Bengali$\rightarrow$English, source-span CKA with the base rises from 0.60 at layer 34 to 0.64 at layer 35 before dropping to 0.43 at layer 36, suggesting that incompatibility is concentrated in the final output-oriented block rather than uniformly across upper layers. As shown in \autoref{tab:cka_summary}, early and intermediate layers preserve broadly shared multilingual representations, but the final decoding layer reorganizes in a target-specific manner. When target languages differ, independently fine-tuned checkpoints no longer occupy a compatible generative geometry, consistent with the degradation observed in One$\rightarrow$Many and bidirectional merging settings.

\subsection{Principal Angle Analysis}
To measure geometric compatibility between independently fine-tuned checkpoints, we compute layer-wise principal angles between their representational subspaces.

\begin{figure}
\centering
\includegraphics[width=1\linewidth]{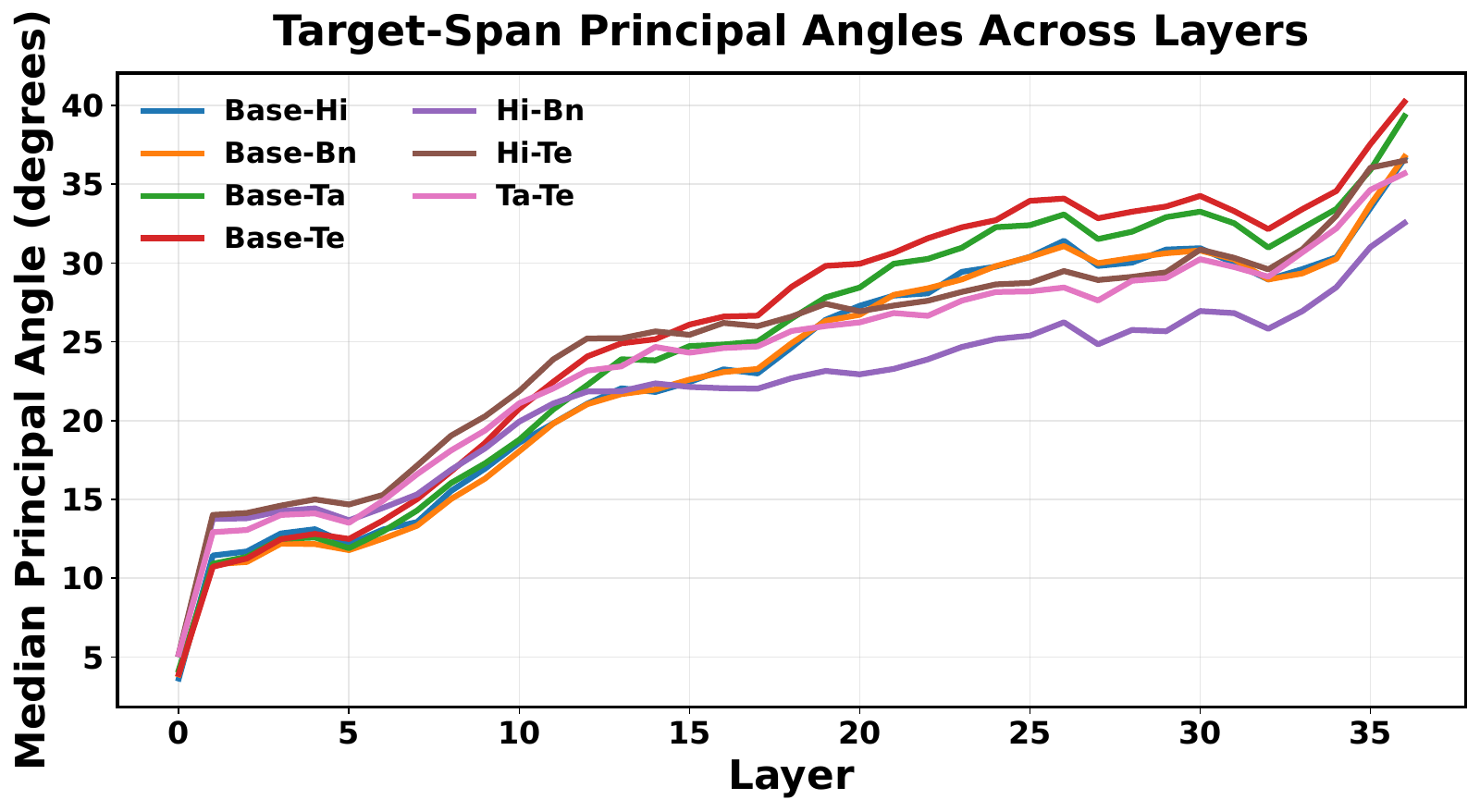}
\caption{Principal angles between representations across models fine-tuned on English$\rightarrow$Indic.}
\label{fig:principal-angles}
\end{figure}

Let $H^{(m)}_l \in \mathbb{R}^{N \times d}$ denote the mean-pooled hidden representations at layer $l$ for model $m$, computed over masked target tokens. After centering $H^{(m)}_l$, we obtain an orthonormal basis $Q^{(m)}_l \in \mathbb{R}^{d \times r}$ from the top-$r$ right singular vectors of the centered representation matrix.

For two models $a$ and $b$, we compute $M_l = Q^{(a)\top}_l Q^{(b)}_l$. If $\sigma_i$ are the singular values of $M_l$, the principal angles are $\theta_i^{(l)} = \arccos(\sigma_i)$. 
We summarize each layer using the median principal angle $\Theta_l^{(a,b)} = \operatorname{median}_i \theta_i^{(l)}$.

\paragraph{Observation.}
Neuron Usage Alignment (NUA) shows that independently fine-tuned bilingual models activate largely overlapping sets of neurons. However, \autoref{fig:principal-angles} shows that, despite this overlap, the dominant representation directions in upper layers diverge substantially for the target span. 

Thus, although the models use similar neurons, their collective activation geometry differs. Merging therefore combines geometrically misaligned subspaces within shared computational units, providing a representation-level explanation for multilingual merging degradation.
\section{Alternatives and Way-forward}

\begin{table*}
\centering
\rowcolors{2}{gray!15}{white}
\resizebox{\textwidth}{!}{%
\begin{tabular}{@{}l*{10}{c}@{}}
\toprule
 & \multicolumn{5}{c}{\textbf{BLEU}} & \multicolumn{5}{c}{\textbf{CHRF}} \\
\textbf{Model} & \textbf{German} & \textbf{French} & \textbf{Arabic} & \textbf{Ukrainian} & \textbf{Average} & \textbf{German} & \textbf{French} & \textbf{Arabic} & \textbf{Ukrainian} & \textbf{Average} \\
\midrule
Base & 23.38 & 34.06 & 10.48 & 9.72 & 19.41 & 53.86 & 59.35 & 36.89 & 36.34 & 46.61 \\
Finetuned - Multilingual & 27.68 & 42.41 & 15.66 & 13.64 & 24.84 & 57.28 & 65.61 & 46.06 & 42.11 & 52.76 \\

\midrule
Finetuned - German & \textbf{29.11} & 1.71 & 0.44 & 0.58 & 7.96 & \textbf{57.57} & 20.95 & 0.68 & 0.84 & 20.01 \\
Finetuned - French & 1.48 & \textbf{41.46} & 0.41 & 0.56 & 10.98 & 20.86 & \textbf{64.79} & 0.67 & 0.81 & 21.78 \\
Finetuned - Arabic & 1.29 & 0.59 & \textbf{15.44} & 0.35 & 4.42 & 4.37 & 1.45 & \textbf{46.70} & 0.65 & 13.29 \\
Finetuned - Ukrainian & 0.84 & 0.74 & 0.47 & \textbf{13.29} & 3.83 & 1.08 & 1.10 & 0.62 & \textbf{42.21} & 11.25 \\
\midrule
\multicolumn{11}{c}{\textbf{\textit{Merged Models}}} \\
\midrule
Task Arithmetic & 24.42 & 38.47 & 12.37 & 4.34 & 19.90 & 54.59 & 63.38 & 40.85 & 28.53 & 46.84 \\
TIES & 21.62 & 35.94 & 5.52 & 2.03 & 16.28 & 50.91 & 60.78 & 22.18 & 22.73 & 39.15 \\
DARE & 25.47 & 38.39 & 10.99 & 2.30 & 19.29 & 54.44 & 62.62 & 36.71 & 25.06 & 44.71 \\
SCE-Merging & 24.82 & 38.04 & 12.69 & 7.98 & 20.89 & 54.92 & 63.28 & 42.31 & 34.79 & 48.83 \\
\bottomrule
\end{tabular}%
}
\caption{BLEU and CHRF scores for different models across languages in the English$\rightarrow$XX translation direction.}
\label{tab:results-en-x}
\end{table*}

\subsection{Effect of Stronger Base Translation Performance}

A possible limitation of the Indic experiments is that the pretrained model has weak zero-shot translation performance in English$\rightarrow$Indic directions. The observed degradation could therefore reflect an unusually difficult setting rather than a general limitation of weight-space merging. To test this, we extend the evaluation to English$\rightarrow$German, English$\rightarrow$French, English$\rightarrow$Arabic, and English$\rightarrow$Ukrainian, where the base model has substantially stronger translation performance. 
As shown in \autoref{tab:results-en-x}, stronger base-model performance reduces the severity of merging failure. Unlike in the English$\rightarrow$Indic setting, the merged checkpoints generally remain usable, and the best merged model slightly outperforms the base on average. Thus, although a stronger base model mitigates catastrophic collapse, most task-specific improvements are still not preserved after merging.
These results suggest that weak base translation performance is not the sole cause of the degradation observed in English$\rightarrow$Indic merging.
Further, the multilingual fine-tuned model performs competitively with the individual models. This indicates that a single checkpoint can effectively handle all four languages and that poor merging performance is not caused by insufficient model capacity.

\begin{table*}
\centering
\rowcolors{2}{gray!15}{white}
\resizebox{\textwidth}{!}{%
\begin{tabular}{@{}l*{10}{c}@{}}
\toprule
 & \multicolumn{5}{c}{\textbf{BLEU}} & \multicolumn{5}{c}{\textbf{CHRF}} \\
\textbf{Model} & \textbf{Hindi} & \textbf{Bengali} & \textbf{Tamil} & \textbf{Telugu} & \textbf{Average} & \textbf{Hindi} & \textbf{Bengali} & \textbf{Tamil} & \textbf{Telugu} & \textbf{Average} \\

\midrule
Fine-tuned SFT & 29.23 & 15.45 & 12.61 & 15.59 & 18.22 & 54.47 & 50.13 & 51.75 & 50.17 & 51.63 \\
Fine-tuned LoRA & 28.38 & 14.95 & 13.18 & 15.32 & 17.96 & 54.04 & 50.13 & 52.73 & 50.86 & 51.94 \\
\midrule
\multicolumn{11}{c}{\textbf{\textit{Merged LoRAs}}} \\
\midrule
Linear & 0.00 & 0.03 & 0.03 & 0.08 & 0.04 & 0.04 & 0.03 & 0.04 & 0.06 & 0.04 \\
TIES & 0.02 & 0.03 & 0.05 & 0.07 & 0.04 & 0.43 & 0.24 & 0.53 & 0.42 & 0.41 \\
DARE & 0.01 & 0.09 & 0.06 & 0.19 & 0.09 & 0.24 & 0.21 & 0.37 & 0.33 & 0.29 \\
\bottomrule
\end{tabular}%
}
\caption{BLEU and CHRF scores for LoRA-based models in the English$\rightarrow$Indic direction.}
\label{tab:results-en-indic-lora}
\end{table*}

\subsection{Adapter Swapping Rather than Merging}
\label{sec:lora-alternative}

One practical response to the poor performance of weight-space merging is to avoid merging full checkpoints and instead maintain language-specific adapters~\cite{zhao-etal-2025-adamergex,dmonte2026improvingtrainingefficiencyreducing}. Prior work suggests that LoRA can be favorable for multilingual tuning, although full-parameter fine-tuning may remain stronger when separately trained language-specific models are considered \citep{chen-etal-2024-monolingual}. At the same time, several studies report that LoRA is not effective for language fine-tuning \cite{biderman2024lora,math12193149,song-etal-2026-small}. More recent work indicates that this gap can be reduced, or even reversed, when adapters are given sufficient capacity through higher rank, careful hyperparameter tuning, and broader module coverage such as feed-forward or all-linear placement \cite{aggarwal-etal-2024-maple,kunz-2025-train}.

\paragraph{Language-specific adapter training.}
For a fair comparison, we train LoRA and full SFT models using the same number of training examples and the same number of epochs. We validate over eight LoRA configurations and select the setting with the lowest validation loss before training the final adapter on the full training data. Table~\ref{tab:results-en-indic-lora} shows that, when adapters are trained from the beginning with sufficient capacity, language-specific LoRA provides competitive performance compared to SFT.

This result should not be interpreted as evidence that LoRA solves the merging problem. Rather, it shows that adapter \emph{swapping} can preserve language-specific performance without requiring full model copies. From a deployment perspective, this is attractive because the base model can be shared while only the adapter weights vary across languages.
In \autoref{tab:results-en-indic-lora}, merged LoRA adapters exhibit near-zero performance. This is consistent with recent evidence that LoRA merging is substantially more challenging \citep{zhang-zhou-2025-unraveling}, especially when LoRA has a higher rank \cite{li-etal-2026-two}. More specialized adapter-merging methods may reduce this degradation.
\paragraph{Serving overhead.}
Adapter swapping also introduces inference-side costs. A serving system must select, load, cache, and route adapters, especially when competitive adapters use high rank and all-linear placement \citep{aggarwal-etal-2024-maple}. Multi-adapter serving systems such as S-LoRA are motivated by these challenges, including heterogeneous batching, adapter memory management, and CPU--GPU movement of adapter weights \citep{MLSYS2024_906419cd}. Recent systems work further reports overheads from dynamic adapter switching and fragmented CUDA kernel calls \citep{kong2024loraswitchboostingefficiencydynamic}. Thus, LoRA reduces storage and enables modular serving, but it does not provide the same abstraction as a single consolidated checkpoint.

\paragraph{Relation to full-checkpoint merging.}
The setting studied in this paper is post-hoc consolidation: independently trained full-parameter bilingual checkpoints already exist, and the goal is to combine them without access to the original training data. LoRA swapping is relevant when adapters are trained from the start and the serving stack supports dynamic routing. In our runs, LoRA training also takes approximately $1.6\times$ the time of full SFT despite its memory and storage advantages, consistent with observations that high-rank LoRA can be slower than full SFT in some regimes \citep{ko2025loraslowerthink}. Moreover, more complex LoRA variants do not always improve target-language learning over standard LoRA \citep{wijewardhana2026loraempiricalstudyeffectiveness}. The comparison in \autoref{tab:results-en-indic-lora} is therefore not entirely fair, since LoRA involves extra validation steps and longer runtime. In summary, LoRA is a practical deployment alternative when the system can retain and route separate adapters, but it assumes access to LoRA-tuned checkpoints.

\begin{table*}
\centering
\rowcolors{2}{gray!15}{white}
\resizebox{\textwidth}{!}{%
\begin{tabular}{@{}ll*{10}{c}@{}}
\toprule
\textbf{Patch Type} & \textbf{$k$}
& \multicolumn{5}{c}{\textbf{BLEU}}
& \multicolumn{5}{c}{\textbf{CHRF}} \\
\cmidrule(lr){3-7}
\cmidrule(lr){8-12}
&
& \textbf{Hindi} & \textbf{Bengali} & \textbf{Tamil} & \textbf{Telugu} & \textbf{Avg.}
& \textbf{Hindi} & \textbf{Bengali} & \textbf{Tamil} & \textbf{Telugu} & \textbf{Avg.} \\
\midrule
Fine-tuned - Individual
& -
& 29.23 & 15.45 & 12.61 & 15.59 & 18.22
& 54.47 & 50.13 & 51.75 & 50.17 & 51.63 \\
Multilingual f/t: En$\rightarrow$Indic & - & 23.59 & 10.75 & 5.05 & 5.25 & 11.16 & 49.59 & 43.71 & 35.28 & 28.75 & 39.33 \\

SCE-Merging & 0
& 9.88 & 3.34 & 0.74 & 0.50 & 3.61
& 34.08 & 28.46 & 19.23 & 13.19 & 23.74 \\ \midrule

Last & 1
& 11.25 & 4.07 & 1.03 & 0.58 & 4.23
& 36.12 & 31.88 & 23.53 & 15.29 & 26.71 \\

Last & 2
& 13.17 & 4.87 & 1.63 & 0.84 & 5.13
& 39.04 & 34.98 & 27.54 & 18.15 & 29.93 \\

Last & 3
& 15.18 & 6.29 & 1.97 & 1.42 & 6.22
& 41.90 & 38.25 & 30.08 & 21.19 & 32.85 \\

Last & 4
& 16.39 & 7.09 & 2.53 & 1.97 & 6.99
& 43.36 & 40.13 & 32.90 & 24.01 & 35.10 \\

Last & 5
& \textbf{16.22} & \textbf{7.33} & \textbf{2.98} & \textbf{2.27} & \textbf{7.20}
& \textbf{43.34} & \textbf{41.13} & \textbf{35.08} & \textbf{26.13} & \textbf{36.42} \\ \midrule

Random & 5
& 11.42 & 4.05 & 0.77 & 0.57 & 4.20
& 35.64 & 30.19 & 19.75 & 14.22 & 24.95 \\
\bottomrule
\end{tabular}
}
\caption{
Layer replacement after selecting the best default SCE-Merging checkpoint on En$\rightarrow$Indic direction. Last-$k$ replaces the final $k$ target-language transformer layers, while Random-5 replaces five randomly selected layers excluding the final five.
}
\label{tab:sce-last-vs-random-layer-replacement}
\end{table*}

\subsection{Implications for Model Merging in Multilingual Setups}

Although not directly discussed previously, our findings have indirect connections to existing literature. \citet{pmlr-v280-qu25a} show that weight-space interpolation can attenuate input-induced features through depth, allowing input-independent components to dominate and degrading performance. In our experiments, merging selectively suppresses language-specific generative features, especially those concentrated in upper transformer layers after fine-tuning. 

Our neuron-level analysis further shows that fine-tuning redistributes language specialization across layers and increases divergence in higher blocks. This connects to Git Re-Basin \citep{ainsworth2023git}, which argues that successful merging requires models to occupy the same basin up to permutation symmetries. When independently fine-tuned translation models develop misaligned upper-layer generative structures, weight averaging combines incompatible subspaces, producing the asymmetries and retention failures observed in our experiments.

Finally, recent work suggests that multilingual LLMs separate language-agnostic semantic circuits from language-specific output-control circuits, and that code-switching can arise when the language-specific pathway is weakened \cite{xiao2026how}. Our results align with this view in the merging setting: failures are stronger when target languages differ, suggesting that shared semantic structure remains relatively compatible, while language-specific generation mechanisms are more fragile under weight-space fusion.

\subsection{Way Forward}

We do not propose a complete multilingual merging algorithm in this work. Instead, we outline directions suggested by our observations. 

\paragraph{Mixture-of-Experts over the final layers.}
A natural direction is to use a Mixture-of-Experts-like architecture over the final transformer layers, but with deterministic language-level hard routing. In this design, the lower and middle layers are merged into a shared multilingual trunk, while the final few transformer layers are retained as target-language-specific experts. At inference time, the target language specified in the translation instruction selects the corresponding upper-layer expert. 

\begin{figure}
\centering
\includegraphics[width=1\linewidth]{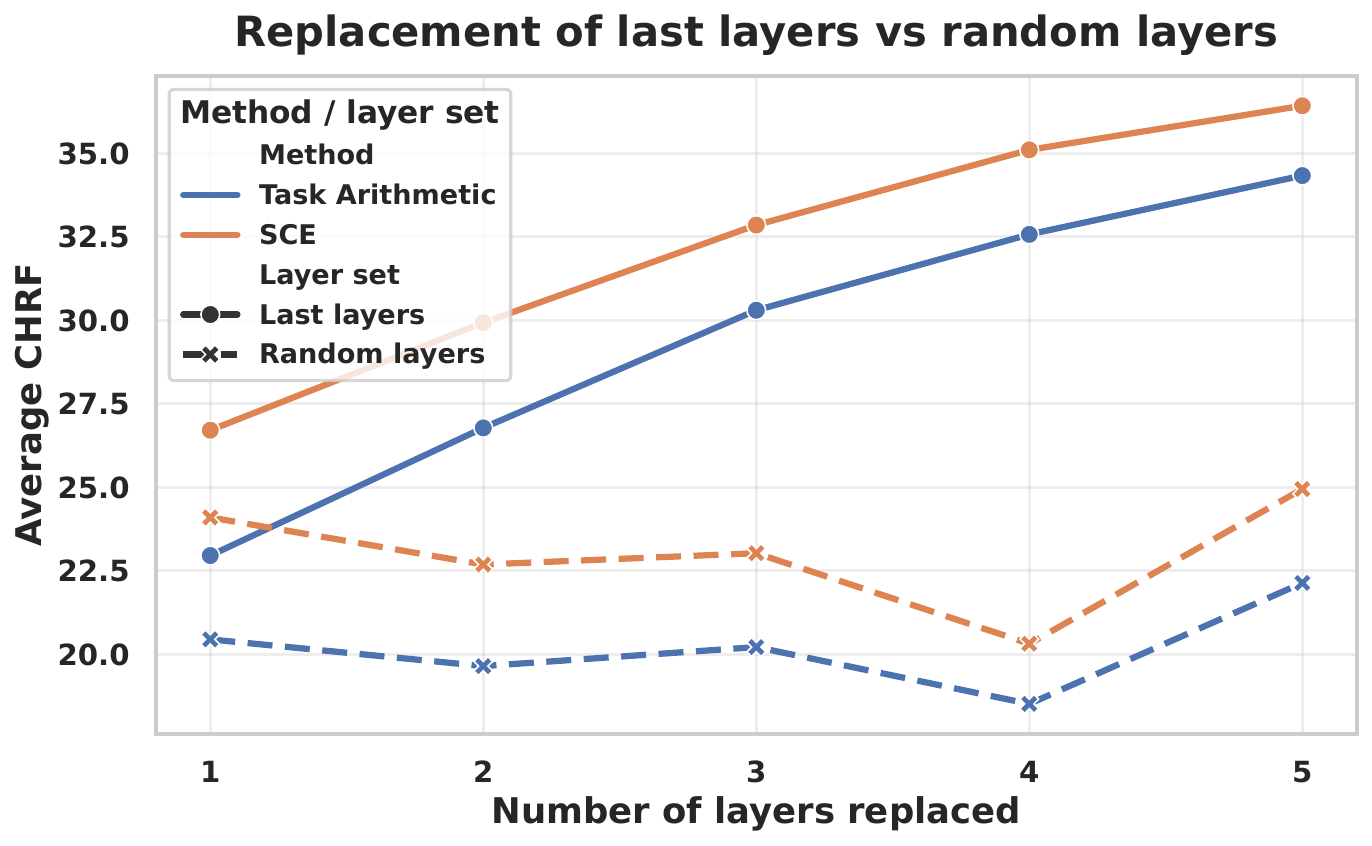}
\caption{Effect of replacing Last-$k$ and Random-$k$ layers with language-specific layers on English$\rightarrow$Indic translation.}
\label{fig:last-layer-replacement}
\end{figure}

The layer replacement experiments provide preliminary support for
investigating this direction. After selecting the best default merged checkpoint, we replace its final $k$ transformer layers with the corresponding layers from the target-language fine-tuned model. As shown in \autoref{tab:sce-last-vs-random-layer-replacement}, replacing the final layers consistently improves SCE-Merging, with larger improvements as $k$ increases. By contrast, replacing five randomly selected non-final layers gives substantially weaker gains. The same pattern is visible in \autoref{fig:last-layer-replacement}, where average CHRF improves steadily under final-layer replacement but remains much lower under random-layer replacement. We find similar observations for Task Arithmetic, and report them in \autoref{tab:ta-last-vs-random-layer-replacement}. We focus on these two methods because they had the highest scores before replacement.
A hard-routed last-layer expert architecture would therefore avoid averaging the most misaligned layers, while still sharing most of the model through the merged lower and middle blocks. This provides a practical compromise between full model merging and maintaining one complete model per target language. However, hard routing, while suitable for our study, may not be suitable for non-translation multilingual setups where the target and source languages may not be obvious. In such cases, training a router \cite{zhou-etal-2025-mergeme} can be considered for soft routing.

\paragraph{Mixing Different Languages during Fine-tuning}
The results in \autoref{tab:results-indic-en} and
\autoref{tab:results-en-indic} show that multilingual fine-tuning is
effective. In both directions, the jointly fine-tuned model substantially
improves over the base, confirming that a single checkpoint can support
multiple language pairs. Unlike post-hoc merging, the multilingual model
assumes joint access to all training data and compute. Thus, poor merging
performance reflects the difficulty of combining independently specialized
checkpoints rather than limited model capacity. Practitioners may retain
some data from related languages during pair-specific fine-tuning
\cite{NEURIPS2019_fa7cdfad,arnal2026efficientrltrainingllms} to preserve
cross-language compatibility and avoid over-specialization.

\section{Conclusion}

We presented a systematic study of weight-space model merging for multilingual machine translation, focusing on the post-hoc consolidation setting in which independently fine-tuned bilingual checkpoints must be combined without access to the original training data. Across shared-target, shared-source, and bidirectional merging configurations, we find a strong directional asymmetry: merging is substantially more effective when models share a target language, but degrades sharply when target-side generation spaces differ.

Our analysis shows that independently fine-tuned models activate largely overlapping neuron sets, but their late-layer representational geometry becomes increasingly misaligned. These findings clarify why standard weight-space merging methods, which are effective in many multitask settings, struggle in multilingual MT. Practically, our results suggest that future multilingual consolidation methods should either preserve cross-language compatibility during fine-tuning or avoid averaging the most target-specific late layers. Overall, our work identifies target-side geometric misalignment as a central obstacle to post-hoc merging of multilingual generative models.


\section*{Acknowledgments}

Baban Gain acknowledges the Prime Minister's Research Fellowship (PMRF), Ministry of Education, Government of India, for supporting this research under Grant No.~2703563. The authors gratefully acknowledge the grant titled ``COIL-D: Centre of Indian Language Data", sponsored by Meity, Govt. of India (Grant No.-11 (14)/2018-HCC(TDIL)Voll-III-Part (21)) and ``GCP Credit Award: Gemma Academic Program" for their partial support. 

\bibliography{tacl2021,anthology-1,anthology-2}

\begin{thebibliography}{68}
\expandafter\ifx\csname natexlab\endcsname\relax\def\natexlab#1{#1}\fi

\bibitem[{Aggarwal et~al.(2024)Aggarwal, Sathe, Watts, and Sitaram}]{aggarwal-etal-2024-maple}
Divyanshu Aggarwal, Ashutosh Sathe, Ishaan Watts, and Sunayana Sitaram. 2024.
\newblock \href {https://doi.org/10.18653/v1/2024.findings-acl.881} {{MAPLE}: Multilingual evaluation of parameter efficient finetuning of large language models}.
\newblock In \emph{Findings of the Association for Computational Linguistics: ACL 2024}, pages 14824--14867, Bangkok, Thailand. Association for Computational Linguistics.

\bibitem[{Aharoni et~al.(2019)Aharoni, Johnson, and Firat}]{aharoni-etal-2019-massively}
Roee Aharoni, Melvin Johnson, and Orhan Firat. 2019.
\newblock \href {https://doi.org/10.18653/v1/N19-1388} {Massively multilingual neural machine translation}.
\newblock In \emph{Proceedings of the 2019 Conference of the North {A}merican Chapter of the Association for Computational Linguistics: Human Language Technologies, Volume 1 (Long and Short Papers)}, pages 3874--3884, Minneapolis, Minnesota. Association for Computational Linguistics.

\bibitem[{Ainsworth et~al.(2023)Ainsworth, Hayase, and Srinivasa}]{ainsworth2023git}
Samuel Ainsworth, Jonathan Hayase, and Siddhartha Srinivasa. 2023.
\newblock \href {https://openreview.net/forum?id=CQsmMYmlP5T} {Git re-basin: Merging models modulo permutation symmetries}.
\newblock In \emph{The Eleventh International Conference on Learning Representations}.

\bibitem[{Arivazhagan et~al.(2019)Arivazhagan, Bapna, Firat, Lepikhin, Johnson, Krikun, Chen, Cao, Foster, Cherry, Macherey, Chen, and Wu}]{arivazhagan2019massivelymultilingualneuralmachine}
Naveen Arivazhagan, Ankur Bapna, Orhan Firat, Dmitry Lepikhin, Melvin Johnson, Maxim Krikun, Mia~Xu Chen, Yuan Cao, George Foster, Colin Cherry, Wolfgang Macherey, Zhifeng Chen, and Yonghui Wu. 2019.
\newblock \href {http://arxiv.org/abs/1907.05019} {Massively multilingual neural machine translation in the wild: Findings and challenges}.

\bibitem[{Arnal et~al.(2026)Arnal, Cabannes, Cohen, Kempe, and Munos}]{arnal2026efficientrltrainingllms}
Charles Arnal, Vivien Cabannes, Taco Cohen, Julia Kempe, and Remi Munos. 2026.
\newblock \href {http://arxiv.org/abs/2604.08706} {Efficient rl training for llms with experience replay}.

\bibitem[{Bandyopadhyay et~al.(2025)Bandyopadhyay, Bhattacharjee, and Ekbal}]{bandyopadhyay2025thinkingmachinessurveyllm}
Dibyanayan Bandyopadhyay, Soham Bhattacharjee, and Asif Ekbal. 2025.
\newblock \href {http://arxiv.org/abs/2503.10814} {Thinking machines: A survey of llm based reasoning strategies}.

\bibitem[{Biderman et~al.(2024)Biderman, Portes, Ortiz, Paul, Greengard, Jennings, King, Havens, Chiley, Frankle, Blakeney, and Cunningham}]{biderman2024lora}
Dan Biderman, Jacob Portes, Jose Javier~Gonzalez Ortiz, Mansheej Paul, Philip Greengard, Connor Jennings, Daniel King, Sam Havens, Vitaliy Chiley, Jonathan Frankle, Cody Blakeney, and John~Patrick Cunningham. 2024.
\newblock \href {https://openreview.net/forum?id=aloEru2qCG} {Lo{RA} learns less and forgets less}.
\newblock \emph{Transactions on Machine Learning Research}.
\newblock Featured Certification.

\bibitem[{Bojar et~al.(2014)Bojar, Buck, Federmann, Haddow, Koehn, Leveling, Monz, Pecina, Post, Saint-Amand, Soricut, Specia, and Tamchyna}]{bojar-etal-2014-findings}
Ond{\v{r}}ej Bojar, Christian Buck, Christian Federmann, Barry Haddow, Philipp Koehn, Johannes Leveling, Christof Monz, Pavel Pecina, Matt Post, Herve Saint-Amand, Radu Soricut, Lucia Specia, and Ale{\v{s}} Tamchyna. 2014.
\newblock \href {https://doi.org/10.3115/v1/W14-3302} {Findings of the 2014 workshop on statistical machine translation}.
\newblock In \emph{Proceedings of the Ninth Workshop on Statistical Machine Translation}, pages 12--58, Baltimore, Maryland, USA. Association for Computational Linguistics.

\bibitem[{Chang et~al.(2024)Chang, Arnett, Tu, and Bergen}]{chang-etal-2024-multilinguality}
Tyler~A. Chang, Catherine Arnett, Zhuowen Tu, and Benjamin~K. Bergen. 2024.
\newblock \href {https://doi.org/10.18653/v1/2024.emnlp-main.236} {When is multilinguality a curse? language modeling for 250 high- and low-resource languages}.
\newblock In \emph{Proceedings of the 2024 Conference on Empirical Methods in Natural Language Processing}, pages 4074--4096, Miami, Florida, USA. Association for Computational Linguistics.

\bibitem[{Chen et~al.(2024)Chen, Ji, Bogoychev, Kutuzov, Haddow, and Heafield}]{chen-etal-2024-monolingual}
Pinzhen Chen, Shaoxiong Ji, Nikolay Bogoychev, Andrey Kutuzov, Barry Haddow, and Kenneth Heafield. 2024.
\newblock \href {https://doi.org/10.18653/v1/2024.findings-eacl.90} {Monolingual or multilingual instruction tuning: Which makes a better alpaca}.
\newblock In \emph{Findings of the Association for Computational Linguistics: EACL 2024}, pages 1347--1356, St. Julian{'}s, Malta. Association for Computational Linguistics.

\bibitem[{Chouhan et~al.(2024)Chouhan, Nath, and Dutta}]{10.1007/978-3-031-78172-81_7}
Sanjay Chouhan, Shubha~Brata Nath, and Aparajita Dutta. 2024.
\newblock \href {https://doi.org/10.1007/978-3-031-78172-8_17} {Hindillm: Large language model for hindi}.
\newblock In \emph{Pattern Recognition: 27th International Conference, ICPR 2024, Kolkata, India, December 1–5, 2024, Proceedings, Part VI}, page 255–270, Berlin, Heidelberg. Springer-Verlag.

\bibitem[{Dmonte et~al.(2026)Dmonte, Gupta, Perry, and Arehart}]{dmonte2026improvingtrainingefficiencyreducing}
Alphaeus Dmonte, Vidhi Gupta, Daniel~J Perry, and Mark Arehart. 2026.
\newblock \href {http://arxiv.org/abs/2601.16127} {Improving training efficiency and reducing maintenance costs via language specific model merging}.

\bibitem[{Fan et~al.(2021)Fan, Bhosale, Schwenk, Ma, El-Kishky, Goyal, Baines, Celebi, Wenzek, Chaudhary, Goyal, Birch, Liptchinsky, Edunov, Auli, and Joulin}]{JMLR:v22:20-1307}
Angela Fan, Shruti Bhosale, Holger Schwenk, Zhiyi Ma, Ahmed El-Kishky, Siddharth Goyal, Mandeep Baines, Onur Celebi, Guillaume Wenzek, Vishrav Chaudhary, Naman Goyal, Tom Birch, Vitaliy Liptchinsky, Sergey Edunov, Michael Auli, and Armand Joulin. 2021.
\newblock \href {http://jmlr.org/papers/v22/20-1307.html} {Beyond english-centric multilingual machine translation}.
\newblock \emph{Journal of Machine Learning Research}, 22(107):1--48.

\bibitem[{Gain et~al.(2026)Gain, Bandyopadhyay, Ekbal, and Singh}]{gain2026bridginglinguisticdividesurvey}
Baban Gain, Dibyanayan Bandyopadhyay, Asif Ekbal, and Trilok~Nath Singh. 2026.
\newblock \href {http://arxiv.org/abs/2504.01919} {Bridging the linguistic divide: A survey on leveraging large language models for machine translation}.

\bibitem[{Gargiulo et~al.(2025)Gargiulo, Crisostomi, Bucarelli, Scardapane, Silvestri, and Rodolà}]{gargiulo2025tasksingularvectorsreducing}
Antonio~Andrea Gargiulo, Donato Crisostomi, Maria~Sofia Bucarelli, Simone Scardapane, Fabrizio Silvestri, and Emanuele Rodolà. 2025.
\newblock \href {http://arxiv.org/abs/2412.00081} {Task singular vectors: Reducing task interference in model merging}.

\bibitem[{Geva et~al.(2021)Geva, Schuster, Berant, and Levy}]{geva-etal-2021-transformer}
Mor Geva, Roei Schuster, Jonathan Berant, and Omer Levy. 2021.
\newblock \href {https://doi.org/10.18653/v1/2021.emnlp-main.446} {Transformer feed-forward layers are key-value memories}.
\newblock In \emph{Proceedings of the 2021 Conference on Empirical Methods in Natural Language Processing}, pages 5484--5495, Online and Punta Cana, Dominican Republic. Association for Computational Linguistics.

\bibitem[{Goddard et~al.(2024)Goddard, Siriwardhana, Ehghaghi, Meyers, Karpukhin, Benedict, McQuade, and Solawetz}]{goddard-etal-2024-arcees}
Charles Goddard, Shamane Siriwardhana, Malikeh Ehghaghi, Luke Meyers, Vladimir Karpukhin, Brian Benedict, Mark McQuade, and Jacob Solawetz. 2024.
\newblock \href {https://doi.org/10.18653/v1/2024.emnlp-industry.36} {Arcee{'}s {M}erge{K}it: A toolkit for merging large language models}.
\newblock In \emph{Proceedings of the 2024 Conference on Empirical Methods in Natural Language Processing: Industry Track}, pages 477--485, Miami, Florida, US. Association for Computational Linguistics.

\bibitem[{Gretton et~al.(2005)Gretton, Bousquet, Smola, and Sch\"{o}lkopf}]{10.1007/11564089_7}
Arthur Gretton, Olivier Bousquet, Alex Smola, and Bernhard Sch\"{o}lkopf. 2005.
\newblock \href {https://doi.org/10.1007/11564089_7} {Measuring statistical dependence with hilbert-schmidt norms}.
\newblock In \emph{Proceedings of the 16th International Conference on Algorithmic Learning Theory}, ALT'05, page 63–77, Berlin, Heidelberg. Springer-Verlag.

\bibitem[{He et~al.(2024)He, Benhaim, Patra, Vaddamanu, Ahuja, Chopra, Chaudhary, Zhao, and Song}]{he2024scalinglawsmultilinguallanguage}
Yifei He, Alon Benhaim, Barun Patra, Praneetha Vaddamanu, Sanchit Ahuja, Parul Chopra, Vishrav Chaudhary, Han Zhao, and Xia Song. 2024.
\newblock \href {http://arxiv.org/abs/2410.12883} {Scaling laws for multilingual language models}.

\bibitem[{He et~al.(2025)He, Zeng, Hu, Yang, Zhang, and Zhao}]{he2025mergebenchbenchmarkmergingdomainspecialized}
Yifei He, Siqi Zeng, Yuzheng Hu, Rui Yang, Tong Zhang, and Han Zhao. 2025.
\newblock \href {http://arxiv.org/abs/2505.10833} {Mergebench: A benchmark for merging domain-specialized llms}.

\bibitem[{Hitit et~al.(2025)Hitit, Girrbach, and Akata}]{hitit2025systematicstudymodelmerging}
Oğuz~Kağan Hitit, Leander Girrbach, and Zeynep Akata. 2025.
\newblock \href {http://arxiv.org/abs/2511.21437} {A systematic study of model merging techniques in large language models}.

\bibitem[{Huang et~al.(2024)Huang, Ye, Chen, He, Yue, and Ouyang}]{huang2024emrmerging}
Chenyu Huang, Peng Ye, Tao Chen, Tong He, Xiangyu Yue, and Wanli Ouyang. 2024.
\newblock \href {https://openreview.net/forum?id=lYdjzx3DYu} {{EMR}-merging: Tuning-free high-performance model merging}.
\newblock In \emph{The Thirty-eighth Annual Conference on Neural Information Processing Systems}.

\bibitem[{Ilharco et~al.(2022)Ilharco, Ribeiro, Wortsman, Gururangan, Schmidt, Hajishirzi, and Farhadi}]{ilharco2022editing}
Gabriel Ilharco, Marco~Tulio Ribeiro, Mitchell Wortsman, Suchin Gururangan, Ludwig Schmidt, Hannaneh Hajishirzi, and Ali Farhadi. 2022.
\newblock Editing models with task arithmetic.
\newblock \emph{arXiv preprint arXiv:2212.04089}.

\bibitem[{Jiang et~al.(2026)Jiang, Wang, Shen, Kim, and Kim}]{10.1145/3747588}
Juyong Jiang, Fan Wang, Jiasi Shen, Sungju Kim, and Sunghun Kim. 2026.
\newblock \href {https://doi.org/10.1145/3747588} {A survey on large language models for code generation}.
\newblock \emph{ACM Trans. Softw. Eng. Methodol.}, 35(2).

\bibitem[{Ko(2025)}]{ko2025loraslowerthink}
Seokmin Ko. 2025.
\newblock \href {http://arxiv.org/abs/2507.08833} {Lora is slower than you think}.

\bibitem[{Kodali et~al.(2025)Kodali, Shivkumar, Joshi, Choudhary, Kumaraguru, and Shrivastava}]{kodali2025adaptingmultilingualmodelscodemixed}
Prashant Kodali, Vaishnavi Shivkumar, Swarang Joshi, Monojit Choudhary, Ponnurangam Kumaraguru, and Manish Shrivastava. 2025.
\newblock \href {http://arxiv.org/abs/2510.19782} {Adapting multilingual models to code-mixed tasks via model merging}.

\bibitem[{Kong et~al.(2024)Kong, Li, Fang, Feng, He, Dong, Wang, Li, Kong, and Liu}]{kong2024loraswitchboostingefficiencydynamic}
Rui Kong, Qiyang Li, Xinyu Fang, Qingtian Feng, Qingfeng He, Yazhu Dong, Weijun Wang, Yuanchun Li, Linghe Kong, and Yunxin Liu. 2024.
\newblock \href {http://arxiv.org/abs/2405.17741} {Lora-switch: Boosting the efficiency of dynamic llm adapters via system-algorithm co-design}.

\bibitem[{Kornblith et~al.(2019)Kornblith, Norouzi, Lee, and Hinton}]{pmlr-v97-kornblith19a}
Simon Kornblith, Mohammad Norouzi, Honglak Lee, and Geoffrey Hinton. 2019.
\newblock \href {https://proceedings.mlr.press/v97/kornblith19a.html} {Similarity of neural network representations revisited}.
\newblock In \emph{Proceedings of the 36th International Conference on Machine Learning}, volume~97 of \emph{Proceedings of Machine Learning Research}, pages 3519--3529. PMLR.

\bibitem[{Kunz(2025)}]{kunz-2025-train}
Jenny Kunz. 2025.
\newblock \href {https://aclanthology.org/2025.nodalida-1.35/} {Train more parameters but mind their placement: {Insights} into language adaptation with {PEFT}}.
\newblock In \emph{Proceedings of the Joint 25th Nordic Conference on Computational Linguistics and 11th Baltic Conference on Human Language Technologies (NoDaLiDa/Baltic-HLT 2025)}, pages 323--330, Tallinn, Estonia. University of Tartu Library.

\bibitem[{Li et~al.(2026)Li, Feng, Ma, Huang, Feng, and Qin}]{li-etal-2026-two}
Zijian Li, Xiachong Feng, Weitao Ma, Yichong Huang, Xiaocheng Feng, and Bing Qin. 2026.
\newblock \href {https://aclanthology.org/2026.findings-acl.1504/} {Two-stage parameter alignment for multi-{L}o{RA} merging in large language models}.
\newblock In \emph{Findings of the {A}ssociation for {C}omputational {L}inguistics: {ACL} 2026}, pages 30079--30093, San Diego, California, United States. Association for Computational Linguistics.

\bibitem[{Loshchilov and Hutter(2019)}]{loshchilov2018decoupled}
Ilya Loshchilov and Frank Hutter. 2019.
\newblock \href {https://openreview.net/forum?id=Bkg6RiCqY7} {Decoupled weight decay regularization}.
\newblock In \emph{International Conference on Learning Representations}.

\bibitem[{Matena and Raffel(2022)}]{matena2022merging}
Michael~S Matena and Colin~A Raffel. 2022.
\newblock Merging models with fisher-weighted averaging.
\newblock \emph{Advances in Neural Information Processing Systems}, 35:17703--17716.

\bibitem[{Meng et~al.(2022)Meng, Bau, Andonian, and Belinkov}]{meng2022locating}
Kevin Meng, David Bau, Alex~J Andonian, and Yonatan Belinkov. 2022.
\newblock \href {https://openreview.net/forum?id=-h6WAS6eE4} {Locating and editing factual associations in {GPT}}.
\newblock In \emph{Advances in Neural Information Processing Systems}.

\bibitem[{Mondal et~al.(2025)Mondal, Sen, Singhania, and Jyothi}]{mondal-etal-2025-language}
Soumen~Kumar Mondal, Sayambhu Sen, Abhishek Singhania, and Preethi Jyothi. 2025.
\newblock \href {https://doi.org/10.18653/v1/2025.insights-1.6} {Language-specific neurons do not facilitate cross-lingual transfer}.
\newblock In \emph{The Sixth Workshop on Insights from Negative Results in NLP}, pages 46--62, Albuquerque, New Mexico. Association for Computational Linguistics.

\bibitem[{OpenAI et~al.(2024)OpenAI, Achiam, Adler, Agarwal, Ahmad, Akkaya, Aleman, Almeida, Altenschmidt, Altman, Anadkat, Avila et~al.}]{openai2024gpt4technicalreport}
OpenAI, Josh Achiam, Steven Adler, Sandhini Agarwal, Lama Ahmad, Ilge Akkaya, Florencia~Leoni Aleman, Diogo Almeida, Janko Altenschmidt, Sam Altman, Shyamal Anadkat, Red Avila, et~al. 2024.
\newblock \href {http://arxiv.org/abs/2303.08774} {Gpt-4 technical report}.

\bibitem[{Papineni et~al.(2002)Papineni, Roukos, Ward, and Zhu}]{papineni-etal-2002-bleu}
Kishore Papineni, Salim Roukos, Todd Ward, and Wei-Jing Zhu. 2002.
\newblock \href {https://doi.org/10.3115/1073083.1073135} {{B}leu: a method for automatic evaluation of machine translation}.
\newblock In \emph{Proceedings of the 40th Annual Meeting of the Association for Computational Linguistics}, pages 311--318, Philadelphia, Pennsylvania, USA. Association for Computational Linguistics.

\bibitem[{Popovi{\'c}(2015)}]{popovic-2015-chrf}
Maja Popovi{\'c}. 2015.
\newblock \href {https://doi.org/10.18653/v1/W15-3049} {chr{F}: character n-gram {F}-score for automatic {MT} evaluation}.
\newblock In \emph{Proceedings of the Tenth Workshop on Statistical Machine Translation}, pages 392--395, Lisbon, Portugal. Association for Computational Linguistics.

\bibitem[{Post(2018)}]{post-2018-call}
Matt Post. 2018.
\newblock \href {https://doi.org/10.18653/v1/W18-6319} {A call for clarity in reporting {BLEU} scores}.
\newblock In \emph{Proceedings of the Third Conference on Machine Translation: Research Papers}, pages 186--191, Brussels, Belgium. Association for Computational Linguistics.

\bibitem[{Qi et~al.(2024)Qi, Li, Wang, Gao, Li, Ye, and Zhou}]{qi2024moreefficientmodelmerging}
Biqing Qi, Fangyuan Li, Zhen Wang, Junqi Gao, Dong Li, Peng Ye, and Bowen Zhou. 2024.
\newblock \href {http://arxiv.org/abs/2412.00054} {Less is more: Efficient model merging with binary task switch}.

\bibitem[{Qu and Horv\'{a}th(2025)}]{pmlr-v280-qu25a}
Xingyu Qu and Samuel Horv\'{a}th. 2025.
\newblock \href {https://proceedings.mlr.press/v280/qu25a.html} {Vanishing feature: Diagnosing model merging and beyond}.
\newblock In \emph{Conference on Parsimony and Learning}, volume 280 of \emph{Proceedings of Machine Learning Research}, pages 1051--1086. PMLR.

\bibitem[{Raihan and Zampieri(2025)}]{raihan-zampieri-2025-tigerllm}
Nishat Raihan and Marcos Zampieri. 2025.
\newblock \href {https://doi.org/10.18653/v1/2025.acl-short.69} {{T}iger{LLM} - a family of {B}angla large language models}.
\newblock In \emph{Proceedings of the 63rd Annual Meeting of the Association for Computational Linguistics (Volume 2: Short Papers)}, pages 887--896, Vienna, Austria. Association for Computational Linguistics.

\bibitem[{Ramesh et~al.(2022)Ramesh, Doddapaneni, Bheemaraj, Jobanputra, AK, Sharma, Sahoo, Diddee, J, Kakwani, Kumar, Pradeep, Nagaraj, Deepak, Raghavan, Kunchukuttan, Kumar, and Khapra}]{ramesh-etal-2022-samanantar}
Gowtham Ramesh, Sumanth Doddapaneni, Aravinth Bheemaraj, Mayank Jobanputra, Raghavan AK, Ajitesh Sharma, Sujit Sahoo, Harshita Diddee, Mahalakshmi J, Divyanshu Kakwani, Navneet Kumar, Aswin Pradeep, Srihari Nagaraj, Kumar Deepak, Vivek Raghavan, Anoop Kunchukuttan, Pratyush Kumar, and Mitesh~Shantadevi Khapra. 2022.
\newblock \href {https://doi.org/10.1162/tacl_a_00452} {Samanantar: The largest publicly available parallel corpora collection for 11 {I}ndic languages}.
\newblock \emph{Transactions of the Association for Computational Linguistics}, 10:145--162.

\bibitem[{Rolnick et~al.(2019)Rolnick, Ahuja, Schwarz, Lillicrap, and Wayne}]{NEURIPS2019_fa7cdfad}
David Rolnick, Arun Ahuja, Jonathan Schwarz, Timothy Lillicrap, and Gregory Wayne. 2019.
\newblock \href {https://proceedings.neurips.cc/paper_files/paper/2019/file/fa7cdfad1a5aaf8370ebeda47a1ff1c3-Paper.pdf} {Experience replay for continual learning}.
\newblock In \emph{Advances in Neural Information Processing Systems}, volume~32. Curran Associates, Inc.

\bibitem[{Rousset et~al.(2025)Rousset, Kakibuchi, Sasaki, and Nomura}]{rousset2025merginglanguagedomainspecific}
Thibault Rousset, Taisei Kakibuchi, Yusuke Sasaki, and Yoshihide Nomura. 2025.
\newblock \href {http://arxiv.org/abs/2502.12001} {Merging language and domain specific models: The impact on technical vocabulary acquisition}.

\bibitem[{Sachan and Neubig(2018)}]{sachan-neubig-2018-parameter}
Devendra Sachan and Graham Neubig. 2018.
\newblock \href {https://doi.org/10.18653/v1/W18-6327} {Parameter sharing methods for multilingual self-attentional translation models}.
\newblock In \emph{Proceedings of the Third Conference on Machine Translation: Research Papers}, pages 261--271, Brussels, Belgium. Association for Computational Linguistics.

\bibitem[{Sel and Hanbay(2024)}]{math12193149}
Ilhami Sel and Davut Hanbay. 2024.
\newblock \href {https://doi.org/10.3390/math12193149} {Efficient adaptation: Enhancing multilingual models for low-resource language translation}.
\newblock \emph{Mathematics}, 12(19).

\bibitem[{Seto et~al.(2025)Seto, Ter~Hoeve, de~Seyssel, and Grangier}]{seto-etal-2025-assessing}
Skyler Seto, Maartje Ter~Hoeve, Maureen de~Seyssel, and David Grangier. 2025.
\newblock \href {https://doi.org/10.18653/v1/2025.findings-emnlp.1236} {Assessing the role of data quality in training bilingual language models}.
\newblock In \emph{Findings of the Association for Computational Linguistics: EMNLP 2025}, pages 22694--22720, Suzhou, China. Association for Computational Linguistics.

\bibitem[{Sheng et~al.(2024)Sheng, Cao, Li, Hooper, Lee, Yang, Chou, Zhu, Zheng, Keutzer, Gonzalez, and Stoica}]{MLSYS2024_906419cd}
Ying Sheng, Shiyi Cao, Dacheng Li, Coleman Hooper, Nicholas Lee, Shuo Yang, Christopher Chou, Banghua Zhu, Lianmin Zheng, Kurt Keutzer, Joseph~E. Gonzalez, and Ion Stoica. 2024.
\newblock \href {https://proceedings.mlsys.org/paper_files/paper/2024/file/906419cd502575b617cc489a1a696a67-Paper-Conference.pdf} {Slora: Scalable serving of thousands of lora adapters}.
\newblock In \emph{Proceedings of Machine Learning and Systems}, volume~6, pages 296--311.

\bibitem[{Skorobogat et~al.(2025)Skorobogat, Roth, and Georgescu}]{skorobogat2025subspaceboostedmodelmerging}
Ronald Skorobogat, Karsten Roth, and Mariana-Iuliana Georgescu. 2025.
\newblock \href {http://arxiv.org/abs/2506.16506} {Subspace-boosted model merging}.

\bibitem[{Song et~al.(2026)Song, Li, Lothritz, Ezzini, Sleem, Gentile, State, Bissyand{\'e}, and Klein}]{song-etal-2026-small}
Yewei Song, Lujun Li, Cedric Lothritz, Saad Ezzini, Lama Sleem, Niccolo' Gentile, Radu State, Tegawend{\'e}~F. Bissyand{\'e}, and Jacques Klein. 2026.
\newblock \href {https://doi.org/10.18653/v1/2026.loresmt-1.1} {Are small language models the silver bullet to low-resource languages machine translation?}
\newblock In \emph{Proceedings for the Ninth Workshop on Technologies for Machine Translation of Low Resource Languages ({L}o{R}es{MT} 2026)}, pages 1--26, Rabat, Morocco. Association for Computational Linguistics.

\bibitem[{Tang et~al.(2024)Tang, Luo, Huang, Zhang, Wang, Zhao, Wei, and Wen}]{tang-etal-2024-language}
Tianyi Tang, Wenyang Luo, Haoyang Huang, Dongdong Zhang, Xiaolei Wang, Xin Zhao, Furu Wei, and Ji-Rong Wen. 2024.
\newblock \href {https://doi.org/10.18653/v1/2024.acl-long.309} {Language-specific neurons: The key to multilingual capabilities in large language models}.
\newblock In \emph{Proceedings of the 62nd Annual Meeting of the Association for Computational Linguistics (Volume 1: Long Papers)}, pages 5701--5715, Bangkok, Thailand. Association for Computational Linguistics.

\bibitem[{Tao et~al.(2024)Tao, Zhang, Huang, Ma, Huang, Zhao, and Feng}]{tao-etal-2024-unlocking}
Mingxu Tao, Chen Zhang, Quzhe Huang, Tianyao Ma, Songfang Huang, Dongyan Zhao, and Yansong Feng. 2024.
\newblock \href {https://doi.org/10.18653/v1/2024.findings-emnlp.508} {Unlocking the potential of model merging for low-resource languages}.
\newblock In \emph{Findings of the Association for Computational Linguistics: EMNLP 2024}, pages 8705--8720, Miami, Florida, USA. Association for Computational Linguistics.

\bibitem[{Team et~al.(2022)Team, Costa-jussà, Cross, Çelebi, Elbayad, Heafield, Heffernan, Kalbassi, Lam, Licht, Maillard, Sun, Wang, Wenzek, Youngblood, Akula, Barrault, Gonzalez, Hansanti, Hoffman, Jarrett, Sadagopan, Rowe, Spruit, Tran, Andrews, Ayan, Bhosale, Edunov, Fan, Gao, Goswami, Guzmán, Koehn, Mourachko, Ropers, Saleem, Schwenk, and Wang}]{nllbteam2022languageleftbehindscaling}
NLLB Team, Marta~R. Costa-jussà, James Cross, Onur Çelebi, Maha Elbayad, Kenneth Heafield, Kevin Heffernan, Elahe Kalbassi, Janice Lam, Daniel Licht, Jean Maillard, Anna Sun, Skyler Wang, Guillaume Wenzek, Al~Youngblood, Bapi Akula, Loic Barrault, Gabriel~Mejia Gonzalez, Prangthip Hansanti, John Hoffman, Semarley Jarrett, Kaushik~Ram Sadagopan, Dirk Rowe, Shannon Spruit, Chau Tran, Pierre Andrews, Necip~Fazil Ayan, Shruti Bhosale, Sergey Edunov, Angela Fan, Cynthia Gao, Vedanuj Goswami, Francisco Guzmán, Philipp Koehn, Alexandre Mourachko, Christophe Ropers, Safiyyah Saleem, Holger Schwenk, and Jeff Wang. 2022.
\newblock \href {http://arxiv.org/abs/2207.04672} {No language left behind: Scaling human-centered machine translation}.

\bibitem[{Tiedemann(2012)}]{tiedemann-2012-parallel}
J{\"o}rg Tiedemann. 2012.
\newblock \href {https://aclanthology.org/L12-1246/} {Parallel data, tools and interfaces in {OPUS}}.
\newblock In \emph{Proceedings of the Eighth International Conference on Language Resources and Evaluation ({LREC}'12)}, pages 2214--2218, Istanbul, Turkey. European Language Resources Association (ELRA).

\bibitem[{Voita et~al.(2024)Voita, Ferrando, and Nalmpantis}]{voita-etal-2024-neurons}
Elena Voita, Javier Ferrando, and Christoforos Nalmpantis. 2024.
\newblock \href {https://doi.org/10.18653/v1/2024.findings-acl.75} {Neurons in large language models: Dead, n-gram, positional}.
\newblock In \emph{Findings of the Association for Computational Linguistics: ACL 2024}, pages 1288--1301, Bangkok, Thailand. Association for Computational Linguistics.

\bibitem[{Wan et~al.(2025)Wan, Zhong, Yang, Chen, and Quan}]{wan-etal-2025-fusechat}
Fanqi Wan, Longguang Zhong, Ziyi Yang, Ruijun Chen, and Xiaojun Quan. 2025.
\newblock \href {https://doi.org/10.18653/v1/2025.emnlp-main.1096} {{F}use{C}hat: Knowledge fusion of chat models}.
\newblock In \emph{Proceedings of the 2025 Conference on Empirical Methods in Natural Language Processing}, pages 21618--21642, Suzhou, China. Association for Computational Linguistics.

\bibitem[{Wijewardhana et~al.(2026)Wijewardhana, Reyes, and Ranathunga}]{wijewardhana2026loraempiricalstudyeffectiveness}
Thamali Wijewardhana, Napoleon~H. Reyes, and Surangika Ranathunga. 2026.
\newblock \href {http://arxiv.org/abs/2606.10428} {Which lora? an empirical study on the effectiveness of lora techniques during multilingual instruction tuning}.

\bibitem[{Wortsman et~al.(2022)Wortsman, Ilharco, Gadre, Roelofs, Gontijo-Lopes, Morcos, Namkoong, Farhadi, Carmon, Kornblith et~al.}]{wortsman2022model}
Mitchell Wortsman, Gabriel Ilharco, Samir~Ya Gadre, Rebecca Roelofs, Raphael Gontijo-Lopes, Ari~S Morcos, Hongseok Namkoong, Ali Farhadi, Yair Carmon, Simon Kornblith, et~al. 2022.
\newblock Model soups: averaging weights of multiple fine-tuned models improves accuracy without increasing inference time.
\newblock In \emph{International conference on machine learning}, pages 23965--23998. PMLR.

\bibitem[{Xiao et~al.(2026)Xiao, Huang, Wang, Lin, He, Shen, and Ye}]{xiao2026how}
Yuxin Xiao, Zhen Huang, Wenxiao Wang, Binbin Lin, Xiaofei He, Xu~Shen, and Jieping Ye. 2026.
\newblock \href {https://openreview.net/forum?id=HIXPyQ1aMq} {How do language models speak languages? a case study on unintended code-switching}.

\bibitem[{Yadav et~al.(2023)Yadav, Tam, Choshen, Raffel, and Bansal}]{10.5555/3666122.3666432}
Prateek Yadav, Derek Tam, Leshem Choshen, Colin Raffel, and Mohit Bansal. 2023.
\newblock Ties-merging: resolving interference when merging models.
\newblock In \emph{Proceedings of the 37th International Conference on Neural Information Processing Systems}, NIPS '23, Red Hook, NY, USA. Curran Associates Inc.

\bibitem[{Yang et~al.(2026)Yang, Shen, Guo, Wang, Cao, Zhang, and Tao}]{10.1145/3787849}
Enneng Yang, Li~Shen, Guibing Guo, Xingwei Wang, Xiaochun Cao, Jie Zhang, and Dacheng Tao. 2026.
\newblock \href {https://doi.org/10.1145/3787849} {Model merging in llms, mllms, and beyond: Methods, theories, applications, and opportunities}.
\newblock \emph{ACM Comput. Surv.}, 58(8).

\bibitem[{Yu et~al.(2024)Yu, Yu, Yu, Huang, and Li}]{3692070.3694452}
Le~Yu, Bowen Yu, Haiyang Yu, Fei Huang, and Yongbin Li. 2024.
\newblock Language models are super mario: absorbing abilities from homologous models as a free lunch.
\newblock In \emph{Proceedings of the 41st International Conference on Machine Learning}, ICML'24. JMLR.org.

\bibitem[{Zhang et~al.(2020)Zhang, Williams, Titov, and Sennrich}]{zhang-etal-2020-improving}
Biao Zhang, Philip Williams, Ivan Titov, and Rico Sennrich. 2020.
\newblock \href {https://doi.org/10.18653/v1/2020.acl-main.148} {Improving massively multilingual neural machine translation and zero-shot translation}.
\newblock In \emph{Proceedings of the 58th Annual Meeting of the Association for Computational Linguistics}, pages 1628--1639, Online. Association for Computational Linguistics.

\bibitem[{Zhang and Zhou(2025)}]{zhang-zhou-2025-unraveling}
Haobo Zhang and Jiayu Zhou. 2025.
\newblock \href {https://doi.org/10.18653/v1/2025.acl-long.1284} {Unraveling {L}o{RA} interference: Orthogonal subspaces for robust model merging}.
\newblock In \emph{Proceedings of the 63rd Annual Meeting of the Association for Computational Linguistics (Volume 1: Long Papers)}, pages 26459--26472, Vienna, Austria. Association for Computational Linguistics.

\bibitem[{Zhang et~al.(2026)Zhang, Jin, Meng, Wang, and Tan}]{ZHANG2026131928}
Yang Zhang, Hanlei Jin, Dan Meng, Jun Wang, and Jinghua Tan. 2026.
\newblock \href {https://doi.org/https://doi.org/10.1016/j.neucom.2025.131928} {A comprehensive survey on automatic text summarization with exploration of llm-based methods}.
\newblock \emph{Neurocomputing}, 663:131928.

\bibitem[{Zhao et~al.(2025)Zhao, Zhang, Wang, Kawaguchi, and Bing}]{zhao-etal-2025-adamergex}
Yiran Zhao, Wenxuan Zhang, Huiming Wang, Kenji Kawaguchi, and Lidong Bing. 2025.
\newblock \href {https://doi.org/10.18653/v1/2025.naacl-long.493} {{A}da{M}erge{X}: Cross-lingual transfer with large language models via adaptive adapter merging}.
\newblock In \emph{Proceedings of the 2025 Conference of the Nations of the Americas Chapter of the Association for Computational Linguistics: Human Language Technologies (Volume 1: Long Papers)}, pages 9785--9800, Albuquerque, New Mexico. Association for Computational Linguistics.

\bibitem[{Zheng et~al.(2024)Zheng, Zhang, Zhang, Ye, and Luo}]{zheng-etal-2024-llamafactory}
Yaowei Zheng, Richong Zhang, Junhao Zhang, Yanhan Ye, and Zheyan Luo. 2024.
\newblock \href {https://doi.org/10.18653/v1/2024.acl-demos.38} {{L}lama{F}actory: Unified efficient fine-tuning of 100+ language models}.
\newblock In \emph{Proceedings of the 62nd Annual Meeting of the Association for Computational Linguistics (Volume 3: System Demonstrations)}, pages 400--410, Bangkok, Thailand. Association for Computational Linguistics.

\bibitem[{Zhou et~al.(2025)Zhou, Karamanolakis, Soto, Rumshisky, Kulkarni, Huang, Ai, and Lu}]{zhou-etal-2025-mergeme}
Yuhang Zhou, Giannis Karamanolakis, Victor Soto, Anna Rumshisky, Mayank Kulkarni, Furong Huang, Wei Ai, and Jianhua Lu. 2025.
\newblock \href {https://doi.org/10.18653/v1/2025.naacl-long.117} {{M}erge{ME}: Model merging techniques for homogeneous and heterogeneous {M}o{E}s}.
\newblock In \emph{Proceedings of the 2025 Conference of the Nations of the Americas Chapter of the Association for Computational Linguistics: Human Language Technologies (Volume 1: Long Papers)}, pages 2315--2328, Albuquerque, New Mexico. Association for Computational Linguistics.

\end{thebibliography}
\bibliographystyle{acl_natbib}
\clearpage
\onecolumn

\appendix

\section{Experimental Setup}

All full-model checkpoints are fine-tuned for $3$ epochs with AdamW
\cite{loshchilov2018decoupled}, using a learning rate of $5\times 10^{-5}$,
$\beta=(0.9,0.999)$, $\epsilon=10^{-8}$, an \texttt{inverse\_sqrt}
learning-rate scheduler, and a random seed of $42$. Training is performed on $8$
GPUs with per-device batch size $8$ and gradient accumulation of $16$, resulting
in an effective training batch size of $1024$. Validation is performed with the
FLORES\footnote{\url{https://huggingface.co/datasets/facebook/flores}} dev set
\cite{nllbteam2022languageleftbehindscaling} for the corresponding language, and
testing is performed on the corresponding FLORES devtest split\footnote{We release the model checkpoints at \url{https://huggingface.co/collections/baban/model-merging-for-machine-translation}}.

For the English$\rightarrow$Indic experiments, we train one model for each of
Hindi, Bengali, Tamil, and Telugu. To test whether the observations extend
beyond the Indic setting, we additionally train English$\rightarrow$German,
English$\rightarrow$French, English$\rightarrow$Arabic, and
English$\rightarrow$Ukrainian models. For German and French, we use the WMT14
dataset\footnote{\url{https://huggingface.co/datasets/wmt/wmt14}}
\cite{bojar-etal-2014-findings}; for Arabic and Ukrainian, we use the OPUS-100
corpus\footnote{\url{https://huggingface.co/datasets/Helsinki-NLP/opus-100}}
\cite{zhang-etal-2020-improving,tiedemann-2012-parallel}. For each of these
additional language pairs, we use up to $1$M English-source parallel sentence
pairs. The corresponding FLORES language columns are used to construct
validation and test sets.

We also train LoRA adapters as a parameter-efficient alternative to full-model
fine-tuning. LoRA models use the same base model and supervised fine-tuning
setup, with cutoff length $1024$, \texttt{pure\_bf16} training,
\texttt{inverse\_sqrt} scheduling, AdamW optimization, maximum gradient norm
$1.0$, per-device training batch size $8$, gradient accumulation $16$, and
per-device evaluation batch size $16$. We search LoRA configurations separately
for each language and select the best configuration by validation loss. The
search considers ranks $r\in\{64,128\}$, scaling factors
$\alpha\in\{128,256\}$, learning rates in $\{10^{-4},5\times10^{-5}\}$, and
dropout $0.05$. We evaluate two target-module groups: feed-forward layers only,
consisting of \texttt{gate\_proj}, \texttt{up\_proj}, and
\texttt{down\_proj}; and all linear projection layers, consisting of
\texttt{q\_proj}, \texttt{k\_proj}, \texttt{v\_proj}, \texttt{o\_proj},
\texttt{gate\_proj}, \texttt{up\_proj}, and \texttt{down\_proj}. After selecting
the best LoRA configuration for each language, we train the final LoRA adapter
on the full training set for that language.
We release the trained model checkpoints at
For model merging, we initially used MergeBench
\cite{he2025mergebenchbenchmarkmergingdomainspecialized} for TIES, DARE, and
Task Arithmetic. We later switched to MergeKit
\cite{goddard-etal-2024-arcees}, since it additionally supports SCE-Merging
while retaining support for TIES, DARE, and Task Arithmetic. For TIES, we sweep
$K \in \{0.1, 0.2, \ldots, 0.9\}$ and scaling factors
in $\{0.1, 0.2, \ldots, 2.5\}$. For DARE, we sweep sparsity
in $\{0.1, 0.2, \ldots, 0.9, 0.99\}$ and scaling factors
in $\{0.1, 0.2, \ldots, 0.9\}$. For Task Arithmetic, we vary scaling factors
in $\{0.1, 0.2, \ldots, 1.0\}$. For SCE-Merging, we vary
\texttt{topk} in $\{0.1, 0.2, \ldots, 0.9\}$. For each method and merging
configuration, we select the merged checkpoint with the highest average BLEU
\cite{papineni-etal-2002-bleu,post-2018-call} score over the languages involved.
Additionally, we report CHRF scores \cite{popovic-2015-chrf}.
We select the lowest-entropy fraction $\rho$ = 0.1 of neurons and use a high-percentile activation threshold $\tau$ = 0.8 when computing activation rates. Varying these values changes absolute counts but preserves the overall trends.
\section{Limitations}

The empirical scope of this work is necessarily bounded by the experimental setting considered. Our core analysis uses a single backbone and scale,
Qwen-2.5-3B-Instruct, which allows controlled mechanistic and representation-level
comparisons across fine-tuning and merging conditions. However, merging behavior
may differ for substantially larger models or for different architectural families. To check whether the observed degradation is specific to this backbone, we also conducted a parallel study with Llama-3.2-1B for En$\rightarrow$Indic directions and observed consistent trends in merging
performance degradation. These results are not included in the main paper due to space constraints and to keep the analysis focused, but they provide additional evidence that the reported effects are not specific to Qwen-2.5-3B-Instruct.

A related point concerns the interpretation of language-specific results. The
results should not be read as showing that merging categorically fails for some
languages while succeeding for others. For example, the relatively stronger
French result in \autoref{tab:results-en-x} reflects the particular merged
checkpoint selected for test evaluation. Since checkpoint selection is based on
the best average validation performance, the reported model corresponds to the
configuration with the strongest overall validation score. We observed other
merge configurations with more balanced performance across languages, but their
average BLEU was lower and therefore they were not selected for final test
evaluation. Thus, apparent language-level asymmetries should be interpreted as
properties of the selected validation-optimal merge configuration rather than as
definitive evidence that merging is intrinsically viable for one language and
not another.

A further limitation concerns the interpretation of the layer-replacement
results in \autoref{tab:sce-last-vs-random-layer-replacement}. Replacing
the final transformer layers consistently improves over the unmodified
SCE-Merging checkpoint and performs better than replacing randomly
selected non-final layers. However, even the best layer-replaced model
remains substantially below both the multilingual and individual
fine-tuning baselines. These experiments therefore support the
localization of merging interference in the upper layers, but do not
establish that a hard-routed last-layer architecture would match
conventional fine-tuning. We consequently present this architecture as
a direction for future investigation rather than as a complete solution.

Finally, our conclusions are conditioned on the full-parameter fine-tuning
recipe used in this study. The language-specific checkpoints exhibit
substantial forgetting outside their supervised translation pairs.
Less aggressive optimization, earlier stopping, stronger
regularization, or partial-parameter adaptation might preserve more
cross-language compatibility and consequently improve mergeability.
We do not investigate this specialization--mergeability trade-off.
Similarly, the jointly fine-tuned baselines use the reported common training recipe rather than an exhaustive multilingual
hyperparameter search. A more extensively optimized joint model could
perform better, which would further increase its advantage over
post-hoc merging.
\section{Use of AI Assistance}
\label{appendix:ai-assistance}

We used ChatGPT~\cite{openai2024gpt4technicalreport} to support manuscript
preparation. Specifically, we used it for grammatical corrections, writing
suggestions, and drafting Bash scripts to automate experiments. All suggested
text and scripts were checked by the authors before use, and all experimental
results and conclusions were independently verified by the authors.
\clearpage







\clearpage

\section{CHRF Scores on Bidirectional Merging}
\label{appendix:chrf-birdirectional}

\begin{table*}[t]
\centering
\rowcolors{2}{gray!15}{white}
\resizebox{\textwidth}{!}{%
\begin{tabular}{lcccccccccc}
\toprule
 & \multicolumn{5}{c}{\textbf{En$\rightarrow$XX}} & \multicolumn{5}{c}{\textbf{XX$\rightarrow$En}} \\
\textbf{Model} & \textbf{Hindi} & \textbf{Bengali} & \textbf{Tamil} & \textbf{Telugu} & \textbf{Average} & \textbf{Hindi} & \textbf{Bengali} & \textbf{Tamil} & \textbf{Telugu} & \textbf{Average} \\
\midrule
Base & 27.97 & 23.54 & 24.12 & 15.78 & 22.85 & 43.29 & 39.08 & 21.26 & 24.76 & 32.10 \\
Multilingual f/t: En$\leftrightarrow$Indic & 46.71 & 41.36 & 34.05 & 26.66 & 37.19 & 51.92 & 44.51 & 31.92 & 27.57 & 38.98 \\
Fine-tuned & 54.47 & 50.13 & 51.75 & 50.17 & - & 62.21 & 59.42 & 54.72 & 57.61 & - \\

\midrule
\multicolumn{11}{c}{\textbf{\textit{Merged Models}}} \\
\midrule
Task Arithmetic & 27.52 & 22.57 & 0.77 & 1.68 & 13.14 & 56.64 & 52.24 & 44.76 & 49.80 & 50.86 \\
TIES & 0.81 & 0.57 & 0.79 & 1.68 & 0.96 & 56.85 & 51.91 & 47.01 & 51.82 & 51.90 \\
DARE & 0.83 & 0.59 & 0.79 & 1.70 & 0.98 & 55.61 & 53.01 & 46.65 & 49.92 & 51.30 \\
SCE-Merging & 33.29 & 34.98 & 0.80 & 1.67 & 17.69 & 56.12 & 51.34 & 46.73 & 50.12 & 51.08 \\

\bottomrule
\end{tabular}%
}
\caption{CHRF scores for bidirectional merging.}
\label{tab:bidirectional-chrf}
\end{table*}

Table~\ref{tab:bidirectional-chrf} reports CHRF scores for bidirectional merging across the English$\rightarrow$Indic and Indic$\rightarrow$English directions. The results show a clear directional asymmetry. For English$\rightarrow$Indic, merged checkpoints degrade substantially, with most methods collapsing on Tamil and Telugu. In contrast, for Indic$\rightarrow$English, all merged models achieve strong CHRF scores and consistently outperform the multilingual fine-tuned baseline. This indicates that merging is more effective when the merged checkpoints share a common target language, since the target-side generation distribution remains aligned across language pairs. When the target languages differ, as in English$\rightarrow$Indic, independently fine-tuned checkpoints develop language-specific generation behaviors that are difficult to combine through direct weight-space merging.

\section{Layer Replacement on Task Arithmetic}
\label{appendix:layer-replacement-task-arith}

\begin{table*}
\centering
\rowcolors{2}{gray!15}{white}
\resizebox{\textwidth}{!}{%
\begin{tabular}{@{}ll*{10}{c}@{}}
\toprule
\textbf{Patch Type} & \textbf{$k$}
& \multicolumn{5}{c}{\textbf{BLEU}}
& \multicolumn{5}{c}{\textbf{CHRF}} \\
\cmidrule(lr){3-7}
\cmidrule(lr){8-12}
&
& \textbf{Hindi} & \textbf{Bengali} & \textbf{Tamil} & \textbf{Telugu} & \textbf{Avg.}
& \textbf{Hindi} & \textbf{Bengali} & \textbf{Tamil} & \textbf{Telugu} & \textbf{Avg.} \\
\midrule
Default & 0
& 6.24 & 2.19 & 0.49 & 0.54 & 2.37
& 28.08 & 23.92 & 17.00 & 12.78 & 20.44 \\ \midrule

Last & 1
& 7.93 & 2.61 & 0.71 & 0.51 & 2.94
& 31.23 & 27.32 & 19.48 & 13.82 & 22.96 \\

Last & 2
& 10.14 & 3.51 & 1.05 & 0.94 & 3.91
& 34.84 & 31.22 & 23.88 & 17.19 & 26.78 \\

Last & 3
& 12.14 & 4.37 & 1.33 & 1.41 & 4.81
& 38.54 & 34.26 & 27.82 & 20.58 & 30.30 \\

Last & 4
& 13.25 & 5.35 & 1.66 & 1.81 & 5.52
& 40.46 & 37.14 & 29.66 & 23.03 & 32.57 \\

Last & 5
& \textbf{13.07} & \textbf{5.70} & \textbf{2.24} & \textbf{2.22} & \textbf{5.81}
& \textbf{40.72} & \textbf{38.53} & \textbf{32.40} & \textbf{25.65} & \textbf{34.33} \\ \midrule

Random & 5
& 8.05 & 2.82 & 0.67 & 0.54 & 3.02
& 30.65 & 25.93 & 18.18 & 13.80 & 22.14 \\
\bottomrule
\end{tabular}
}
\caption{
Layer replacement after selecting the best default Task Arithmetic checkpoint. Last-$k$ replaces the final $k$ target-language transformer layers, while Random-5 replaces five randomly selected layers excluding the final five.
}
\label{tab:ta-last-vs-random-layer-replacement}
\end{table*}

Table~\ref{tab:ta-last-vs-random-layer-replacement} analyzes whether replacing selected target-language-specific layers after merging can recover lost performance. Replacing the final transformer layers leads to steady improvements in both BLEU and CHRF. The gains increase as more final layers are restored. In contrast, replacing five randomly selected non-final layers provides only a small improvement over the default checkpoint. This supports the observation that the most severe merging damage is concentrated in the upper transformer blocks, which are closely tied to target-side autoregressive generation.

\section{Layer-wise Language-Specific Neurons}
\label{sec:layerwise-heatmap}

\begin{figure*}
\centering
\textbf{Source Masked Activations}\\[0.6em]

\begin{minipage}{0.28\textwidth}
\centering
\includegraphics[width=\linewidth]{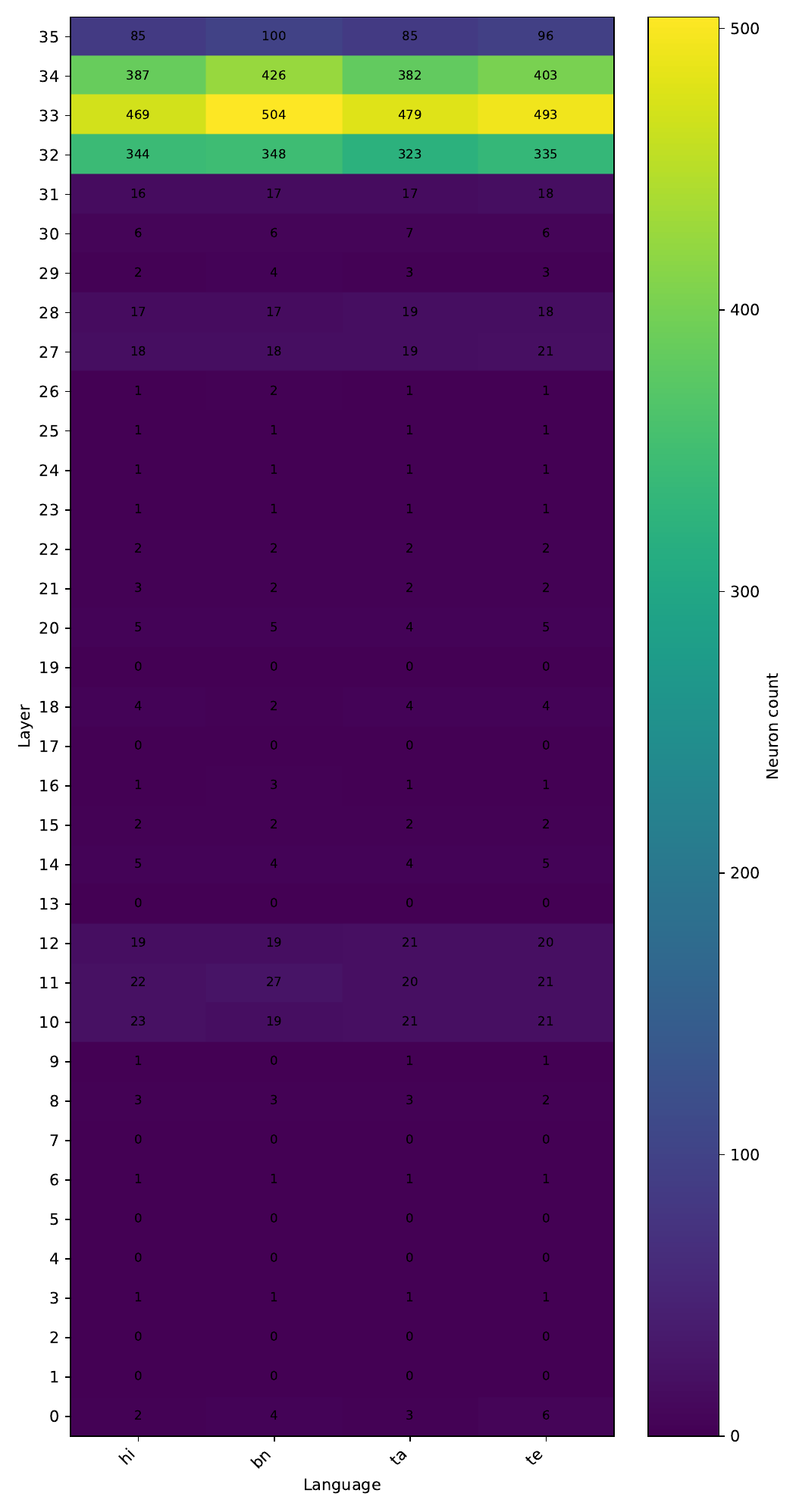}
\\[-0.3em]
{\small English$\rightarrow$Bengali}

\vspace{0.6em}

\includegraphics[width=\linewidth]{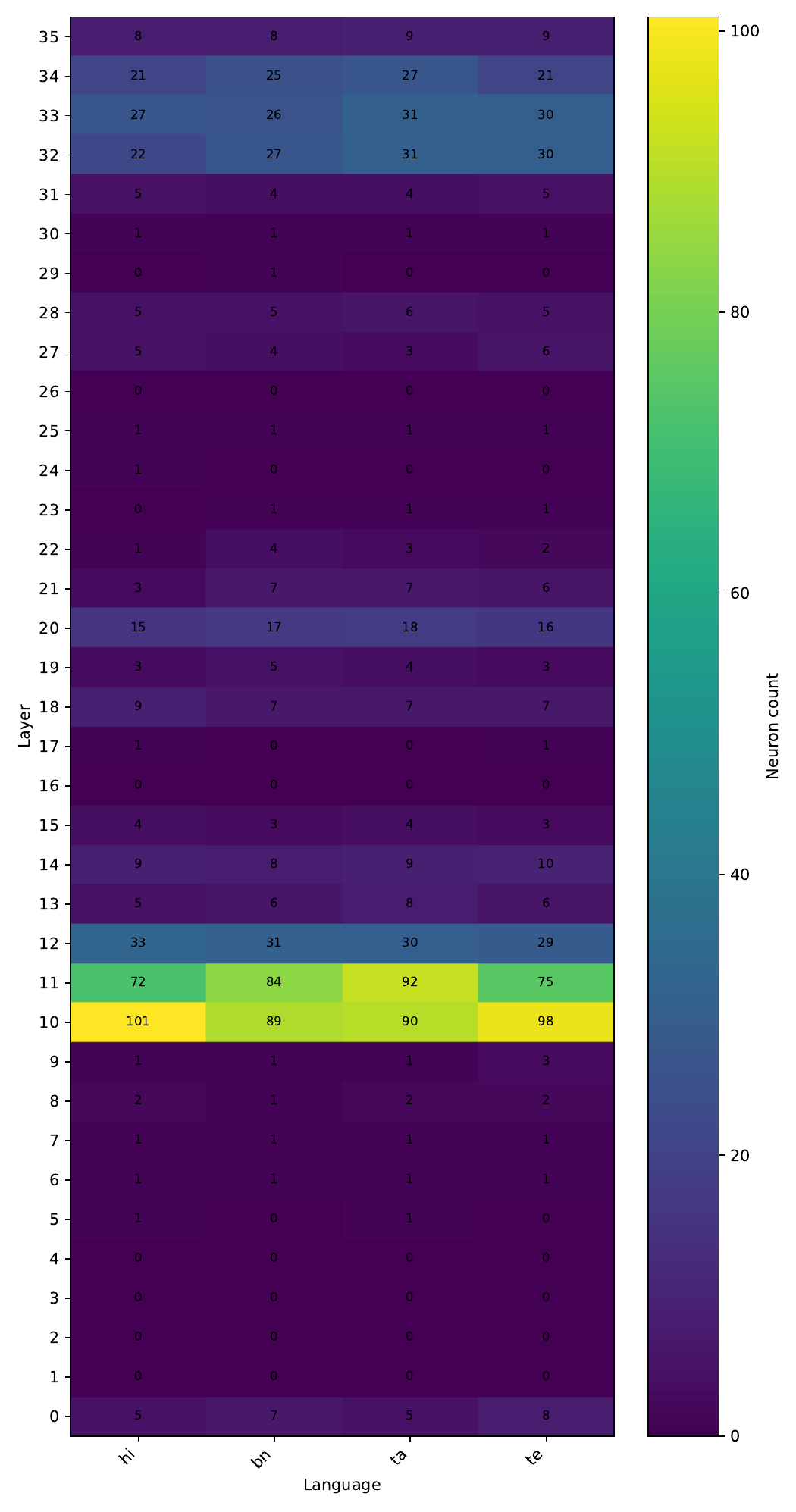}
\\[-0.3em]
{\small English$\rightarrow$Tamil}
\end{minipage}
\hfill
\begin{minipage}{0.36\textwidth}
\centering
\vspace{2.2em}
\includegraphics[width=\linewidth]{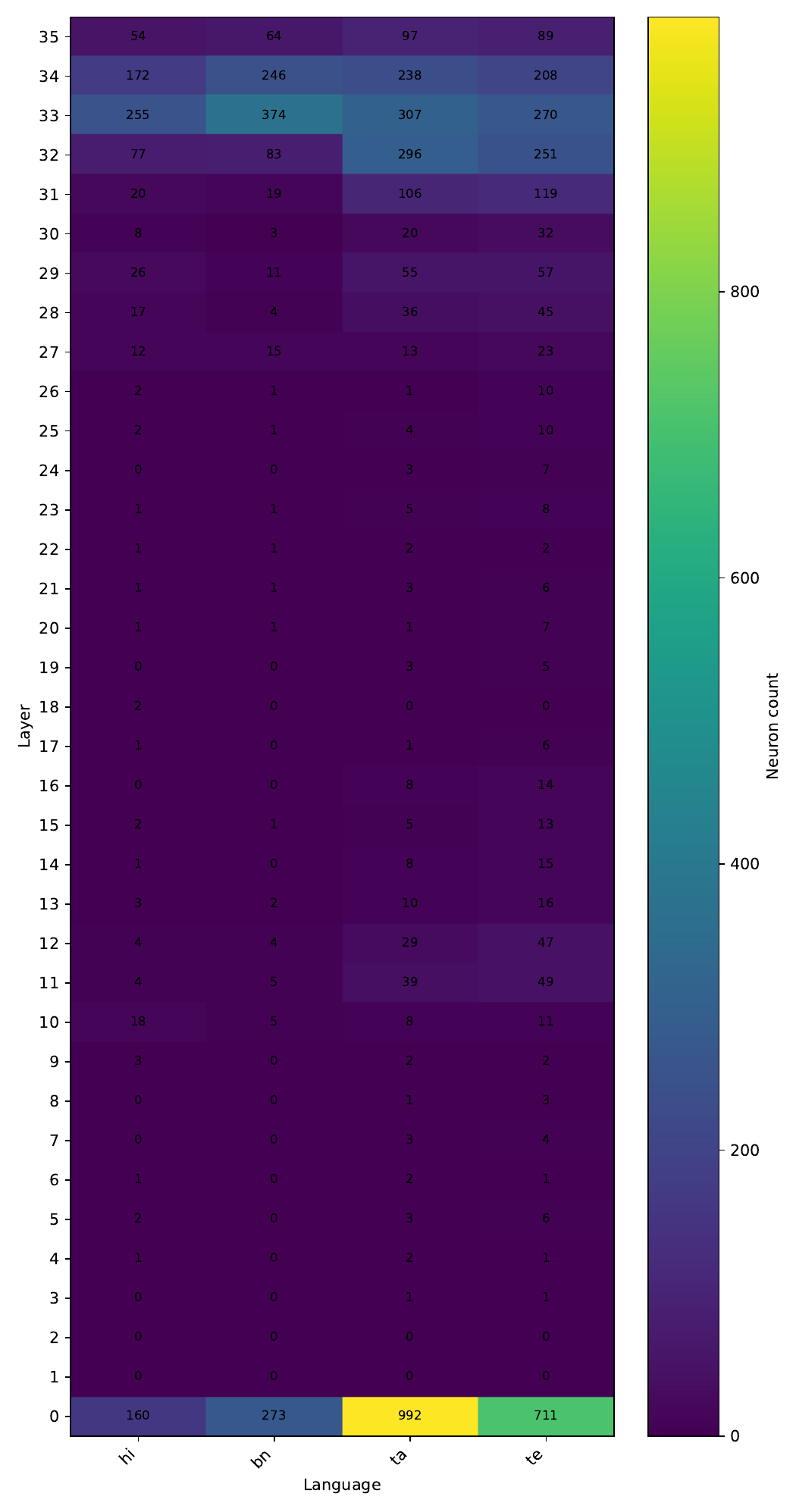}
\\[-0.3em]
{\small \textbf{Instruct}}
\end{minipage}
\hfill
\begin{minipage}{0.28\textwidth}
\centering
\includegraphics[width=\linewidth]{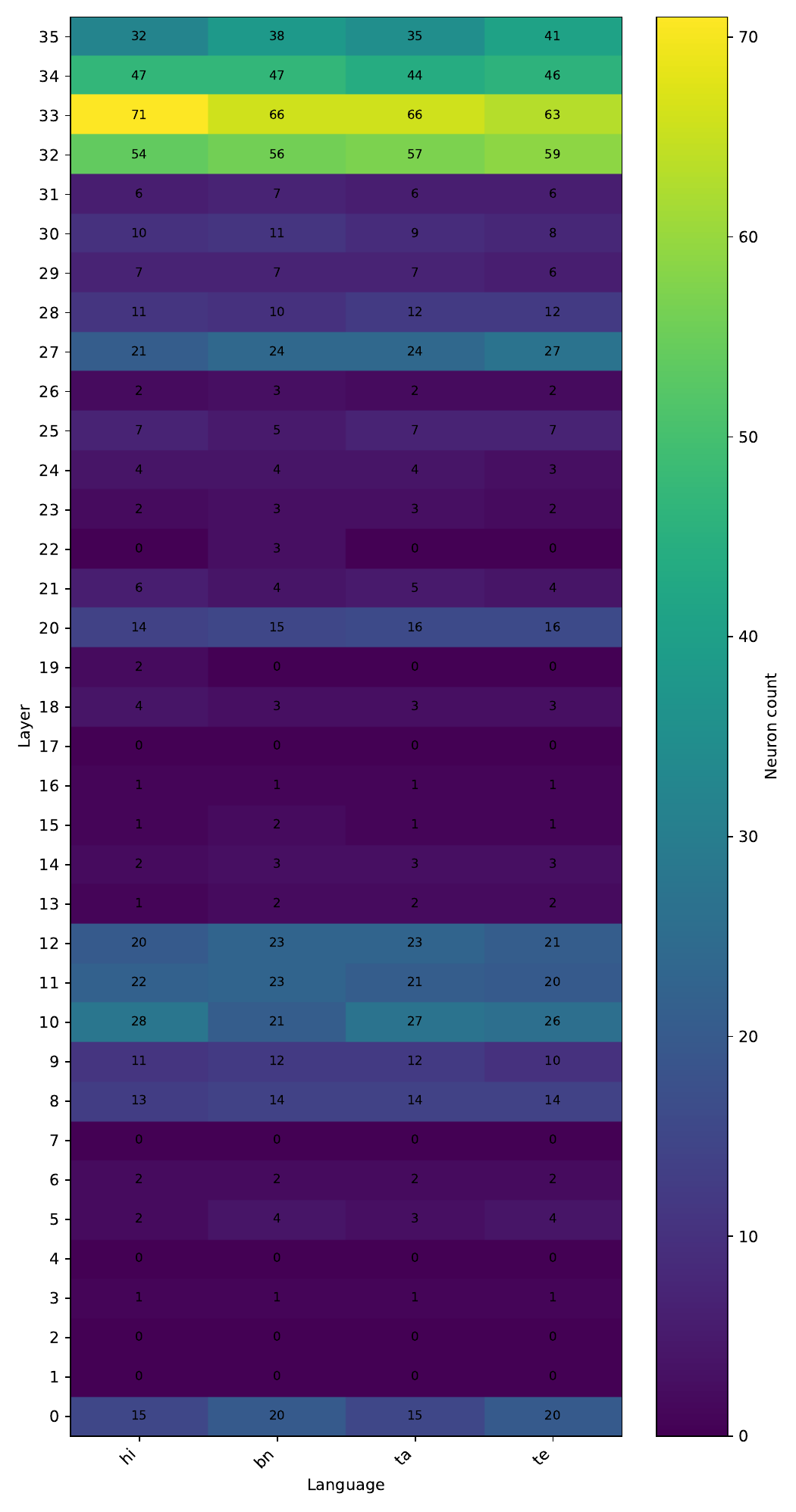}
\\[-0.3em]
{\small English$\rightarrow$Hindi}

\vspace{0.6em}

\includegraphics[width=\linewidth]{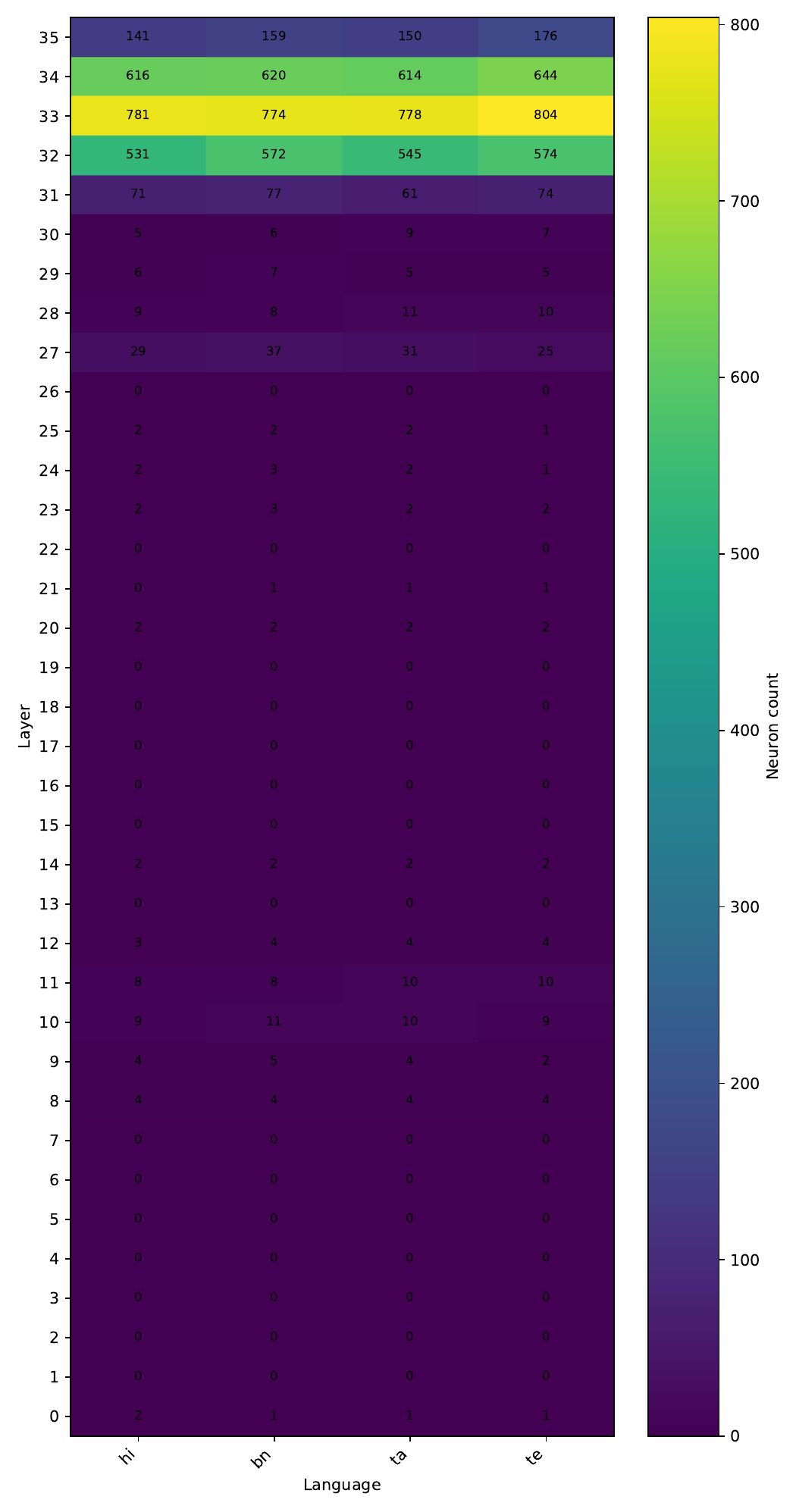}
\\[-0.3em]
{\small English$\rightarrow$Telugu}
\end{minipage}

\caption{Layer-wise neuron counts under source masking for English$\rightarrow$Indic.}
\label{fig:matrix_src_compact_en_indic}
\end{figure*}

\begin{figure*}
\centering
\textbf{Target Masked Activations}\\[0.6em]

\begin{minipage}{0.28\textwidth}
\centering
\includegraphics[width=\linewidth]{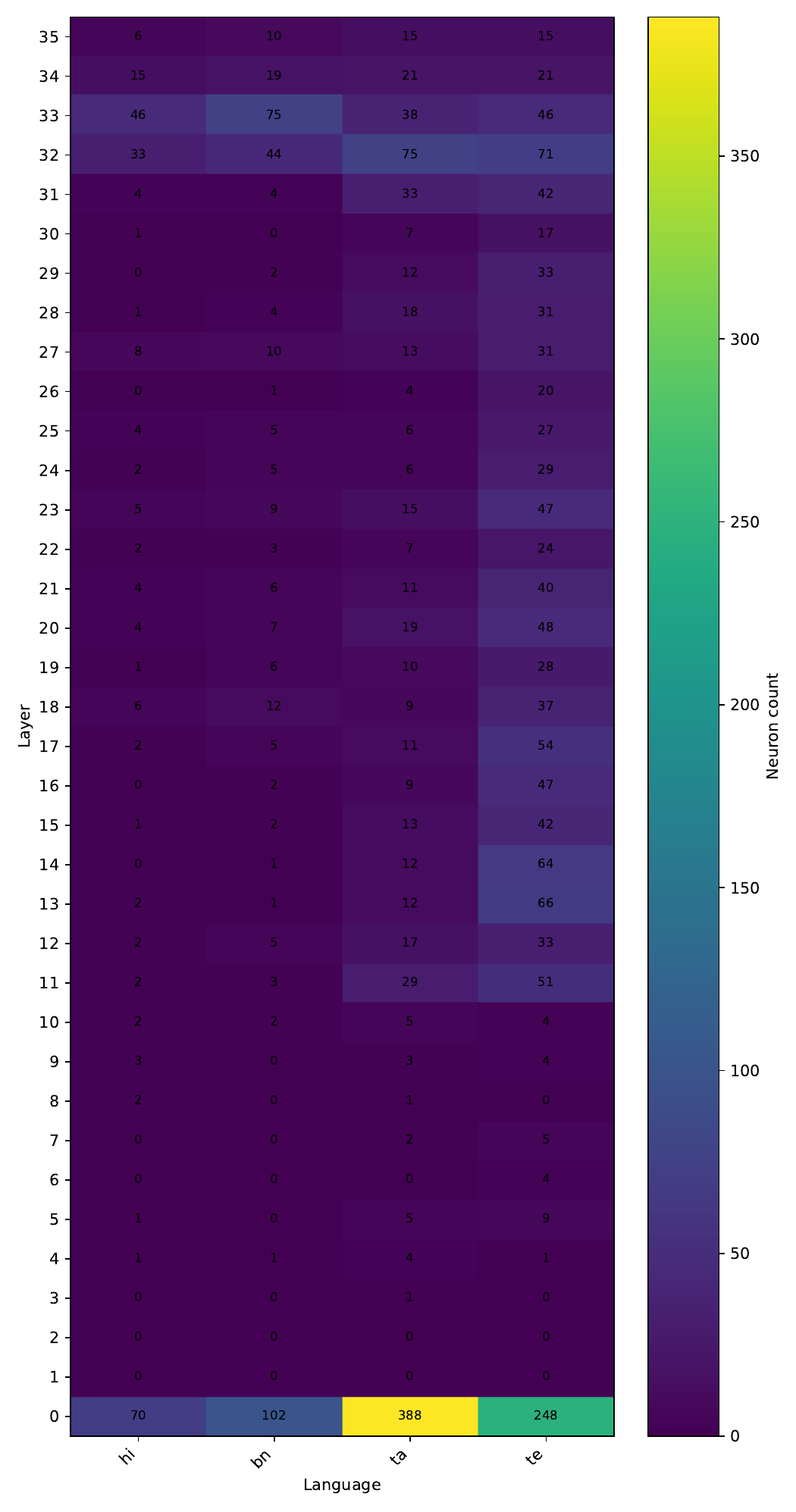}
\\[-0.3em]
{\small English$\rightarrow$Bengali}

\vspace{0.6em}

\includegraphics[width=\linewidth]{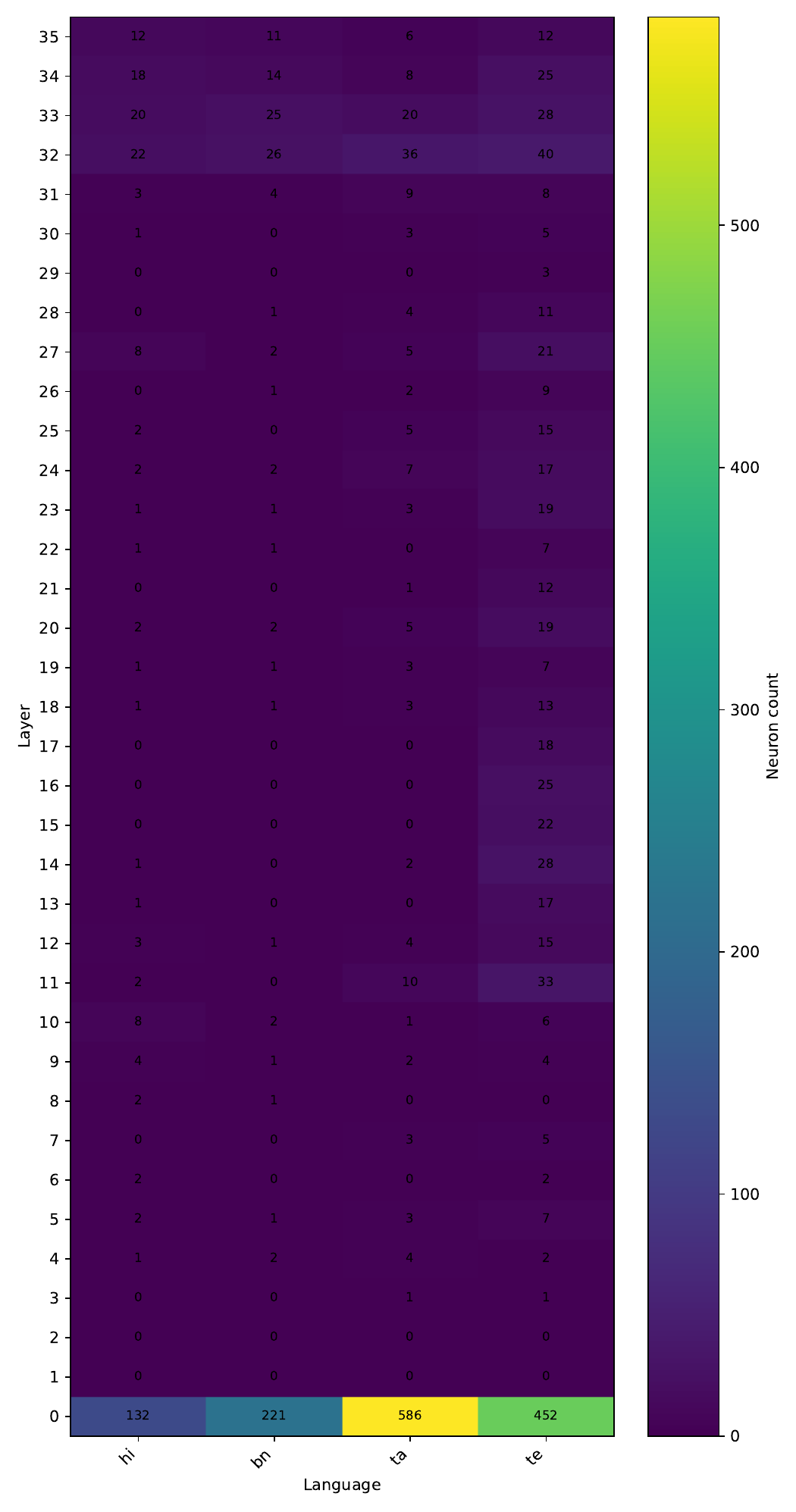}
\\[-0.3em]
{\small English$\rightarrow$Tamil}
\end{minipage}
\hfill
\begin{minipage}{0.36\textwidth}
\centering
\vspace{2.2em}
\includegraphics[width=\linewidth]{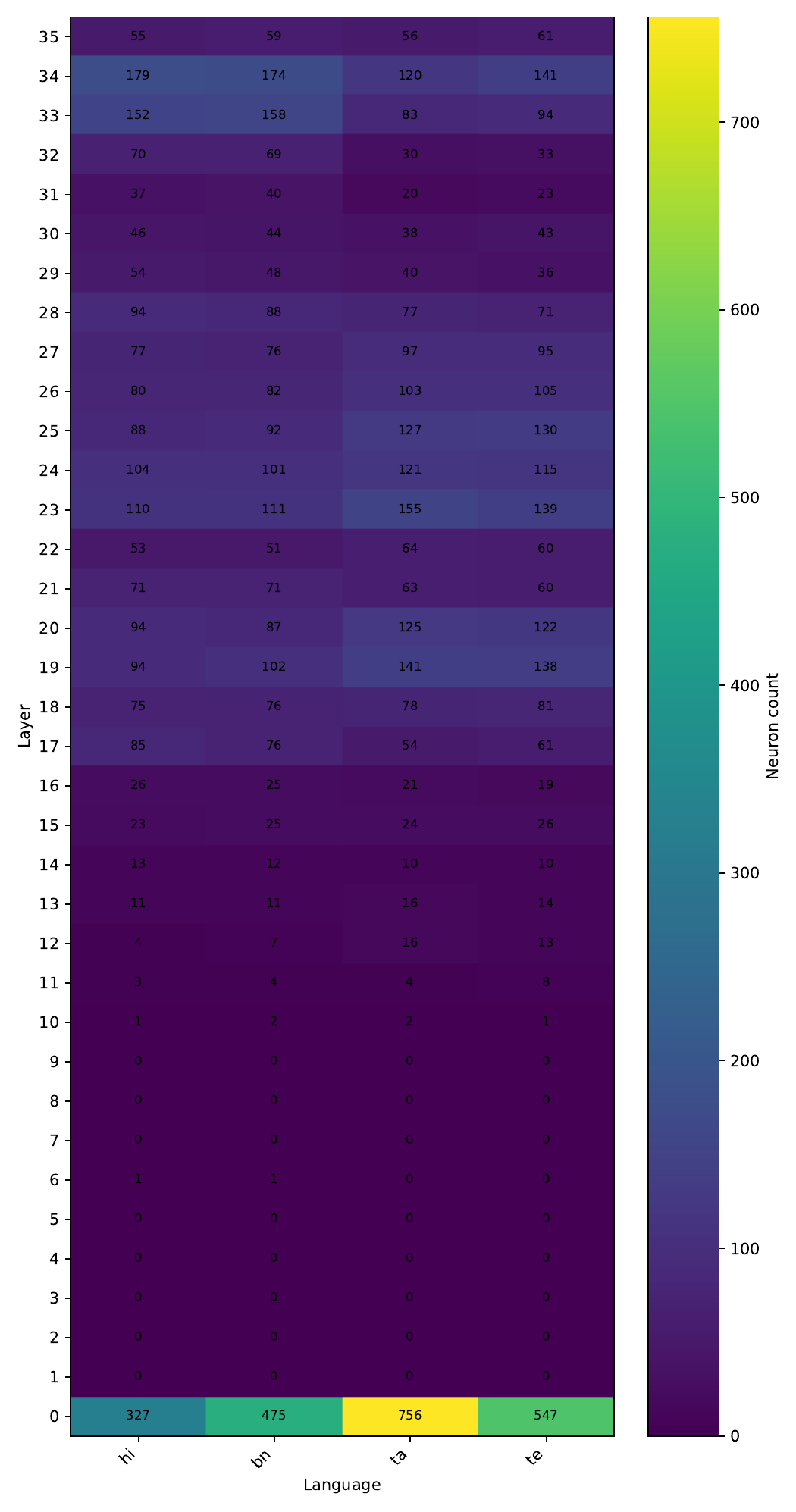}
\\[-0.3em]
{\small \textbf{Instruct}}
\end{minipage}
\hfill
\begin{minipage}{0.28\textwidth}
\centering
\includegraphics[width=\linewidth]{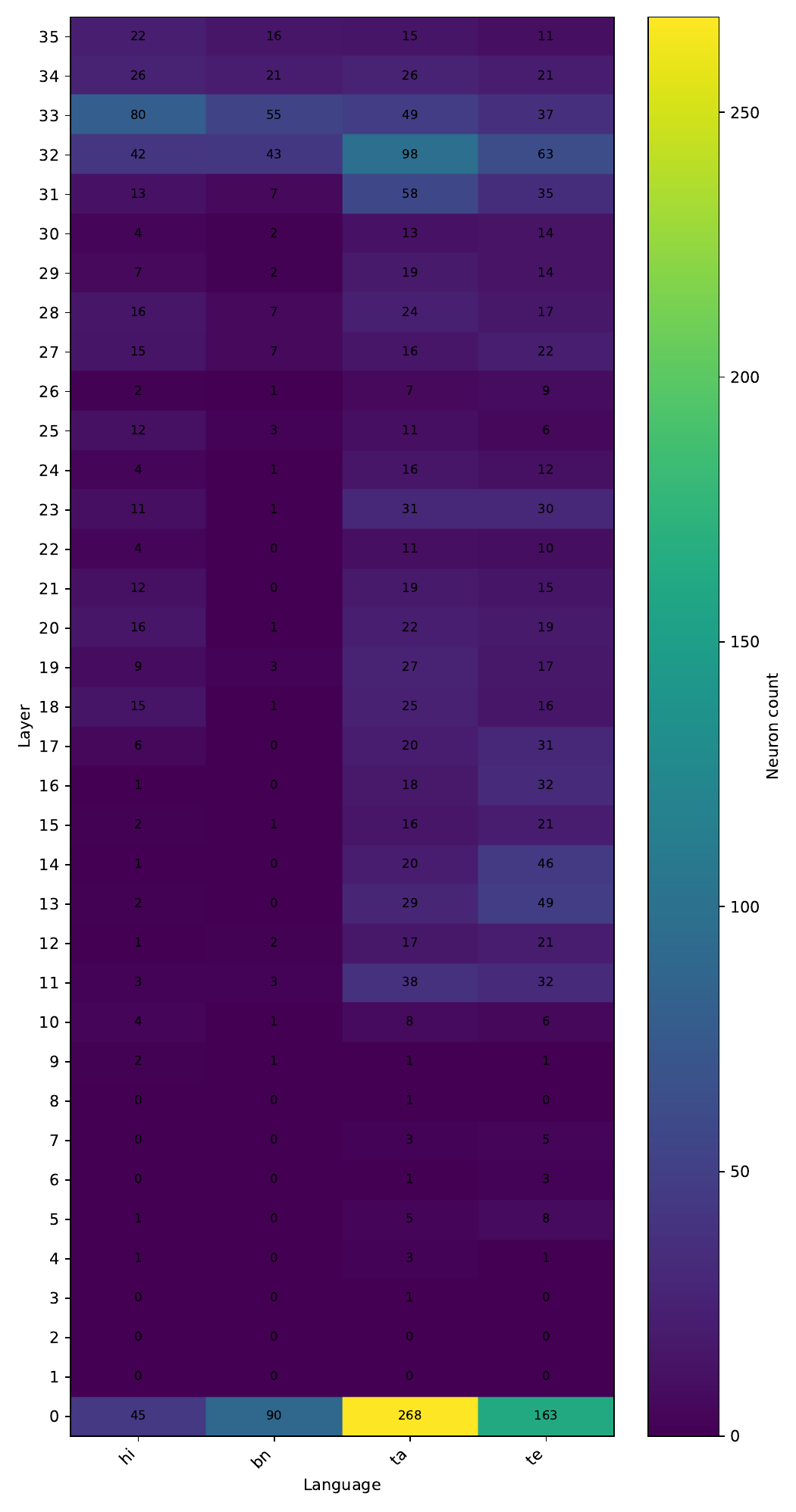}
\\[-0.3em]
{\small English$\rightarrow$Hindi}

\vspace{0.6em}

\includegraphics[width=\linewidth]{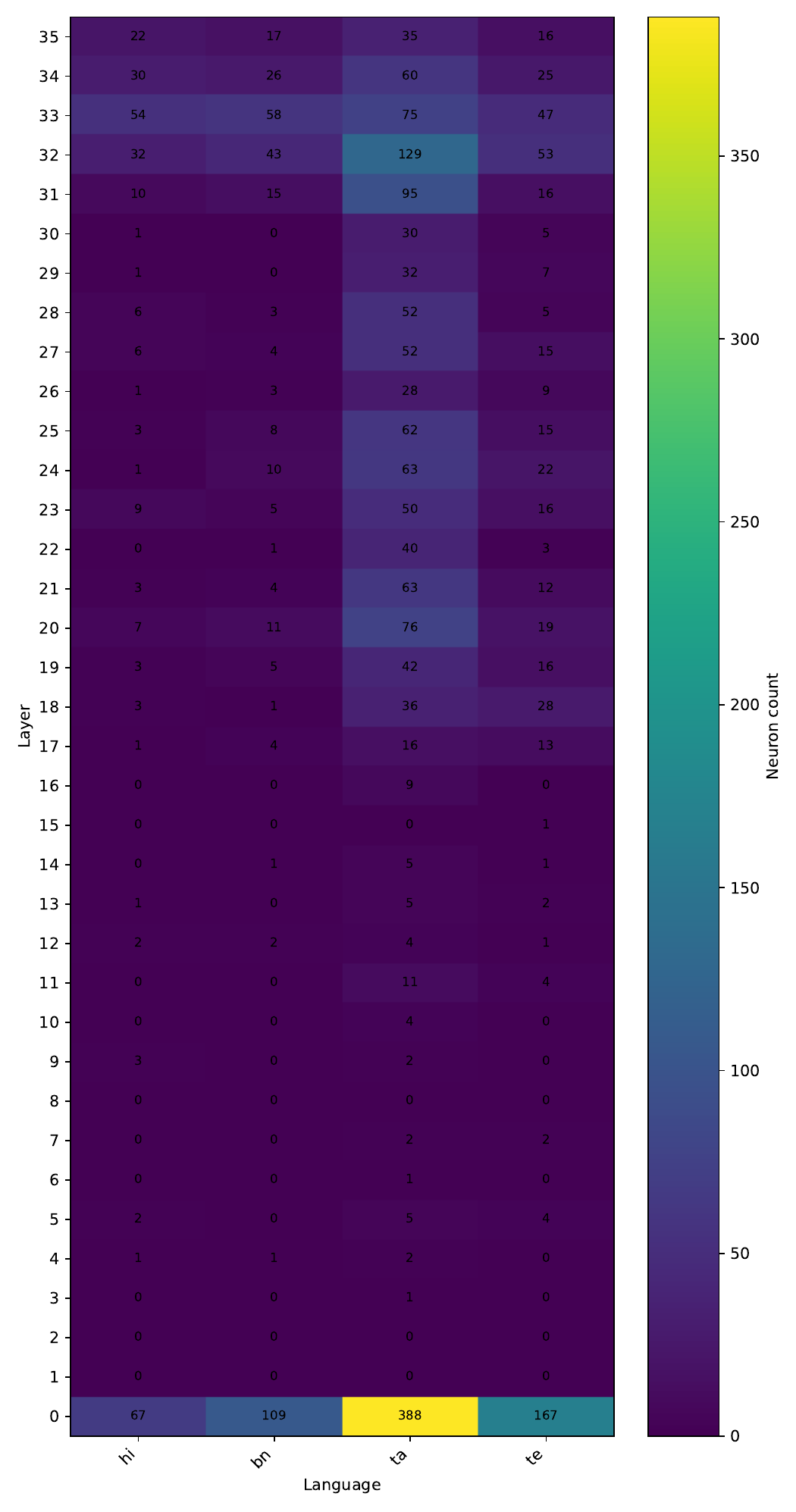}
\\[-0.3em]
{\small English$\rightarrow$Telugu}
\end{minipage}

\caption{Layer-wise neuron counts under target masking for English$\rightarrow$Indic.}
\label{fig:matrix_tgt_compact_en_indic}
\end{figure*}

Language-specific neurons are mostly concentrated in the last layers of the network, especially for the En$\rightarrow$Indic direction, as shown for source and target spans in \autoref{fig:matrix_src_compact_en_indic} and \autoref{fig:matrix_tgt_compact_en_indic}.

\clearpage

\end{document}